\documentclass[runningheads,table]{llncs}

\usepackage{eccv}

\usepackage{eccvabbrv}

\usepackage{graphicx}
\usepackage{booktabs}

\usepackage{caption}

\usepackage[accsupp]{axessibility}  

\usepackage{hyperref}

\usepackage{orcidlink}

\usepackage{amsmath}
\usepackage{amssymb}
\usepackage{tabularx}
\usepackage{tablefootnote}
\usepackage{multirow}
\usepackage{bbm}
\usepackage{comment}

\usepackage{pbox}

\usepackage{pifont}
\newcommand{\cmark}{\ding{51}}%
\newcommand{\xmark}{\ding{55}}%
\newcommand{\gain}[1]{\textcolor[rgb]{0.18, 0.40, 0.86}{#1}}
\newcommand{\dec}[1]{\textcolor[rgb]{0.925, 0.00, 0.545}{#1}}
\newcommand{\gray}[1]{\textcolor[rgb]{0.18, 0.4, 0.46}{#1}}

\newcommand{\captfont}[1]{\fontsize{8}{10}\selectfont{#1 \vspace{-15pt}}}
\newcommand{\cappfont}[1]{\fontsize{8}{10}\selectfont{#1 \vspace{-15pt}}}
\newcommand{\capfont}[1]{\cappfont{#1}}

\usepackage{algorithm}
\usepackage{algorithmicx}
\usepackage{algpseudocode}

\algdef{SE}[DOUNTIL]{Do}{doUntil}{\algorithmicdo}[1]{\algorithmicuntil\ #1}

\usepackage[capitalize]{cleveref}
\crefname{section}{Sec.}{Secs.}
\Crefname{section}{Section}{Sections}
\Crefname{table}{Table}{Tables}
\crefname{table}{Tab.}{Tabs.}

\begin{document}

\title{Weakly Supervised Co-training with Swapping Assignments for Semantic Segmentation}

\titlerunning{CoSA}

\author{
Xinyu Yang\inst{1}\and
Hossein Rahmani\inst{1}\and
Sue Black\inst{2}\and
Bryan M. Williams\inst{1}
}
\authorrunning{Xinyu et al.}

\institute{
Lancaster University, Bailrigg Lancaster, LA1 4YW, UK \and
University of Oxford, Wellington Square Oxford, OX1 2JD, UK \\
\email{\{xyang28,h.rahmani,b.williams6\}@lancaster.ac.uk, sue.black@sjc.ox.ac.uk}
}
\maketitle

\begin{abstract}
Class activation maps (CAMs) are commonly employed in weakly supervised semantic segmentation (WSSS) to produce pseudo-labels. Due to incomplete or excessive class activation, existing studies often resort to offline CAM refinement, introducing additional stages or proposing offline modules. This can cause optimization difficulties for single-stage methods and limit generalizability. In this study, we aim to reduce the observed CAM inconsistency and error to mitigate reliance on refinement processes. We propose an end-to-end WSSS model incorporating guided CAMs, wherein our segmentation model is trained while concurrently optimizing CAMs online. Our method, Co-training with Swapping Assignments (CoSA), leverages a dual-stream framework, where one sub-network learns from the swapped assignments generated by the other. We introduce three techniques in this framework: i) soft perplexity-based regularization to penalize uncertain regions; ii) a threshold-searching approach to dynamically revise the confidence threshold; and iii) contrastive separation to address the coexistence problem. CoSA demonstrates exceptional performance, achieving mIoU of 76.2\% and 51.0\% on VOC and COCO validation datasets, respectively, surpassing existing baselines by a substantial margin. Notably, CoSA is the first single-stage approach to outperform all existing multi-stage methods including those with additional supervision.
Source code is publicly available at \href{https://github.com/youshyee/CoSA}{here}.
  \keywords{Weakly-supervised Learning \and Semantic Segmentation \and CAM}
\end{abstract}

\section{Introduction} \label{sec:intro}
The objective of weakly supervised semantic segmentation (WSSS) is to train a segmentation model without relying on pixel-level labels but on weak and cost-effective annotations, such as image-level classification labels \cite{araslanov2020single,jiang2022l2g,rong2023boundary}, object points \cite{bearman2016s,liu2021one}, and bounding boxes \cite{dai2015boxsup,khoreva2017simple,song2019box,lee2021bbam}.
In particular, image-level classification labels have commonly been employed as weak labels due to the minimal or negligible annotation effort required \cite{ahn2018learning,wang2020self}. With the absence of precise localization information, image-level WSSS often makes use of the coarse localization offered by class activation maps (CAMs) \cite{zhou2016learning}. CAMs pertain to the intermediate outputs derived from a classification network. They visually illustrate the activation regions corresponding to each individual class. Thus, they are often used to generate pseudo masks for training. However, CAMs suffer from i) Inconsistent Activation: CAMs demonstrate variability and lack robustness in accommodating geometric transformations of input images \cite{wang2020self}, resulting in inconsistent activation regions for the same input. ii) Inaccurate Activation: activation region accuracy is often compromised, resulting in incomplete or excessive class activation, only covering discriminative object regions \cite{ahn2019weakly}.
Despite enhanced localization mechanisms in the variants GradCAM \cite{selvaraju2017grad} and GradCAM$^{++}$ \cite{chattopadhay2018grad}, they still struggle to generate satisfactory pseudo-labels for WSSS \cite{wang2020self}. Thus, many WSSS works are dedicated to studying CAM refinement or post-processing \cite{ahn2019weakly,cheng2023out,kweon2023weakly}.

In general, they \cite{ahn2018learning,xu2022multi,du2022weakly,rong2023boundary} comprise three stages: CAM generation, refinement, and segmentation training with pseudo-labels. Multi-stage frameworks are known to be time-consuming and complex as several models must be trained at different stages. In contrast, single-stage models \cite{araslanov2020single,zhang2020reliability,ru2022learning}, which include a unified network of all stages, are more efficient. They are trained to co-optimize the segmentation and classification tasks, but the generated CAMs are not explicitly trained. As a result, they need refinement to produce high-quality pseudo-labels, often leveraging hand-crafted modules, such as CRF in \cite{zhang2020reliability}, PAMR in \cite{araslanov2020single}, PAR in \cite{ru2022learning,ru2023token}. As the refinement modules are predefined and offline, they decouple the CAMs from the primary optimization. When the refined CAMs are employed as segmentation learning objectives, the optimization of the segmentation branch may deviate from that of the classification branch. Hence, it is difficult for single-stage models to optimize the segmentation task while yielding satisfactory CAM pseudo-labels. This optimization difficulty underlies the inferior performance in single-stage approaches compared to multi-stage \cite{xu2022multi,rong2023boundary}. Further, hand-crafted refinement modules require heuristic tuning and empirical changes, thereby limiting their adaptability to novel datasets \cite{araslanov2020single,ru2022learning}. Despite the potential benefits of post-refinement in addressing the aforementioned issues associated with CAMs, which have been extensively discussed in WSSS studies, there has been limited exploration of explicit online optimization for CAMs.

The absence of fully optimized CAMs is an important factor in the indispensability of this refinement.
In this paper, we take a different approach by optimizing CAMs in an end-to-end fashion. We ask a core question: Can we train a model that delivers reliable, consistent and accurate CAMs, which can be applied directly for WSSS without the necessity for subsequent refinements? We show that the answer is \emph{yes}, in two respects:
1) we note that even though CAM is differentiable, it is not robust to variation. As the intermediate output of classification, CAMs are not fully optimized for segmentation purposes since the primary objective is to minimize classification error. This implies that within an optimized network, numerous weight combinations exist that can yield accurate classification outcomes, while generating CAMs of varying qualities. To investigate this, we conduct \emph{oracle experiments}, training a classification model while simultaneously guiding the CAMs with the segmentation ground truth. A noticeable enhancement in quality is observed in guided compared to vanilla CAMs, without compromising classification accuracy. 2) we demonstrate the feasibility of substituting the oracle with segmentation pseudo-labels (SPL) in the context of weak supervision. Consequently, we harness the potential of SPL for WSSS by co-training both CAMs and segmentation through mutual learning.

We explore an effective way to substitute the CAM refinement process, \textit{i.e.} guiding CAMs in an end-to-end fashion. Our method optimizes the CAMs and segmentation prediction simultaneously thanks to the differentiability of CAMs. To achieve this, we adopt a dual-stream framework that includes an online network (ON) and an assignment network (AN), inspired by self-training frameworks \cite{caron2020unsupervised,grill2020bootstrap,yang2023dynamic}. The AN is responsible for producing CAM pseudo-labels (CPL) and segmentation pseudo-labels (SPL) to train the ON. Since CPL and SPL are swapped for supervising segmentation and CAMs, respectively, our method is named \textbf{Co}-training with \textbf{S}wapping \textbf{A}ssignments (CoSA).

The benefit of this end-to-end framework is that it enables us to quantify pseudo-label reliability online, as opposed to {the offline hard pseudo-labels used in existing} methodologies \cite{ahn2018learning,xu2022multi,cheng2023out,rong2023boundary}.
We can then incorporate soft regularization to compensate for CPL uncertainty, where the segmentation loss for different regions is adaptively weighted according to our estimated perplexity map.
In comparison to existing literature, this dynamic learning scheme can exploit the potential of CPL and enhance the final performance, as opposed to performance being constrained by predetermined CPL.
The threshold is a key hyper-parameter for generating the CPL \cite{wang2020self,rong2023boundary,ru2023token}. It not only requires tuning but necessitates dynamic adjustment to align with the model's learning state at various time-steps. CoSA integrates threshold searching to dynamically adapt its learning state, as opposed to the fixed thresholding \cite{du2022weakly,chen2022c,ru2022learning}. This can enhance performance and help to eliminate the laborious manual parameter-tuning task. We further address a common issue with CAMs, known as the coexistence problem, whereby certain class activations display extensive false positive regions that inaccurately merge the objects with their surroundings (\cref{fig:coex}). In response, we introduce a technique to leverage low-level CAMs enriched with object-specific details to contrastively separate those coexistent classes.

The proposed CoSA greatly surpasses existing WSSS methods. Our approach achieves the leading results on VOC and COCO benchmarks, highlighting the contribution of this work:
\textbf{i)} We are the first to propose SPL as a substitute for guiding CAMs. We present compelling evidence of its potential to produce more reliable, consistent and accurate CAMs.
\textbf{ii)} We introduce a dual-stream framework with swapped assignments, which co-optimizes the CAMs and segmentation predictions in an end-to-end fashion.
\textbf{iii)} We address the learning dynamics, proposing two components within our framework: reliability-based adaptive weighting and dynamic thresholding.
\textbf{iv)} We address the CAM coexistence issue, proposing a contrastive separation approach to regularize CAMs, significantly enhancing the results of affected classes.

\section{Related Work}
\noindent\textbf{Multi-Stage WSSS.}Most image-level WSSS work is multi-stage, typically comprising three stages: CAM generation, CAM refinement, and segmentation training.
Some approaches employ heuristic strategies to address incomplete activation regions, such as adversarial erasing \cite{zhang2018adversarial,sun2021ecs,kweon2021unlocking,yoon2022adversarial}, feature map optimization \cite{lee2019ficklenet,chen2022class,chen2022c,chen2023extracting}, self-supervised learning \cite{wang2020self,chen2022self, shimoda2019self}, and contrastive learning \cite{ke2020universal,xie2022c2am,zhou2022regional,cheng2023out}. Some methods focus on post-refining the CAMs by propagating object regions from the seeds to their semantically similar pixels. AffinityNet \cite{ahn2018learning}, for instance, learns pixel-level affinity to enhance CPL. This has motivated other work \cite{ahn2019weakly,fan2020cian,chen2020weakly,li2022towards} that utilize additional networks to generate more accurate CPL.
Other work {studies }optimization given coarse pseudo-labels: \cite{li2022uncertainty} explores uncertainty of noisy labels, \cite{liu2022adaptive} adaptively corrects CPL during early learning, and \cite{rong2023boundary} enhances boundary prediction through co-training.
Since image-level labels alone do not yield satisfactory results, several methods incorporate additional modalities, such as saliency maps \cite{lee2021railroad,li2022towards,zhou2022regional,du2022weakly} or CLIP models  \cite{xu2023self,lin2023clip,xie2022clims}.
Recently, vision transformers \cite{dosovitskiy2021image} have emerged as prominent models for various vision tasks. Several WSSS studies benefit from vision transformers: \cite{gao2021ts} enhances CAMs by incorporating the attention map from ViT;
\cite{xu2022multi} introduces class-specific attention for discriminative object localization;
\cite{lin2023clip} and \cite{xu2023self} leverage multi-modal transformers to enhance performance.

\vspace{3pt}\noindent\textbf{Single-Stage WSSS.}
In contrast, single-stage methods are much more efficient.
They contain a shared backbone with heads for classification and segmentation \cite{araslanov2020single,zhang2020reliability,ru2022learning,ru2023token}. The pipeline involves generating and refining the CAMs, leveraging an offline module, such as PAMR \cite{araslanov2020single}, PAR \cite{ru2022learning}, or CRF \cite{zhang2020reliability}. Subsequently, the refined CPL are used for segmentation.
Single-stage methods exhibit faster speed and a lower memory footprint but are challenging to optimize due to the obfuscation in offline refinement. As a result, they often yield inferior performance compared to multi-stage methods. More recently, with the success of ViT, single-stage WSSS has been greatly advanced. AFA \cite{ru2022learning} proposes learning reliable affinity from attention to refine the CAMs. Similarly, ToCo \cite{ru2023token} mitigates the problem of over-smoothing in vision transformers by contrastively learning from patch tokens and class tokens. The existing works depend heavily on offline refinement of CAMs.
In this study, we further explore the potential of single-stage approaches and showcase the redundancy of offline refinement. We propose an effective alternative for generating consistent, and accurate CAMs in WSSS.

\section{Method}
\subsection{Guiding Class Activation Maps} \label{sec:gcam}
Class activation maps are determined by the feature map $F$ and the weights $W_{\text{fc}}$ for the last \texttt{FC} layer \cite{zhou2016learning}. Let us consider a $C$ classes classification problem:
\begin{equation}\label{eq:cls}
  \small
  \mathcal{L}_{\text{cls}} (Z , Y)\!=\! \frac{-1}{C} \sum_{c=1}^{C}\! \Big[ Y^{c}\log \sigma_Z^c + (1-Y^c)\log \left(1 - \sigma_Z^c\right) \Big],
\end{equation}
where $\sigma_Z^c\triangleq\sigma ({ Z^c})$ represents \texttt{Sigmoid} activation, $Y\triangleq Y_{\text{gt}}$ denotes the one-hot multi-class label, and $Z\triangleq GW_{\text{fc}}^{\top}\!\in\!\mathbb{R}^{C}$ represents the prediction logits, derived from the final \texttt{FC} layer, where $G\!=\!\texttt{Pooling}(F)\!\in\!\mathbb{R}^{D}$ is a spatial pooled feature from $F\!\in\!\mathbb{R}^{HW\times D}$. During training, \cref{eq:cls} is optimized with respect to the learnable parameters $\theta$ in the backbone{.}
When gradients flow backwards from $G$ to $F$, only a fraction of elements in $F$ get optimized, implying that a perturbation in $F$ does not guarantee corresponding response in $G$ due to the spatial pooling, resulting in non-determinism in the feature map $F$. This indeterminate nature can lead to stochasticity of the generated CAMs.

\begin{figure}[t]
  \centering
  \includegraphics[width=1\linewidth]{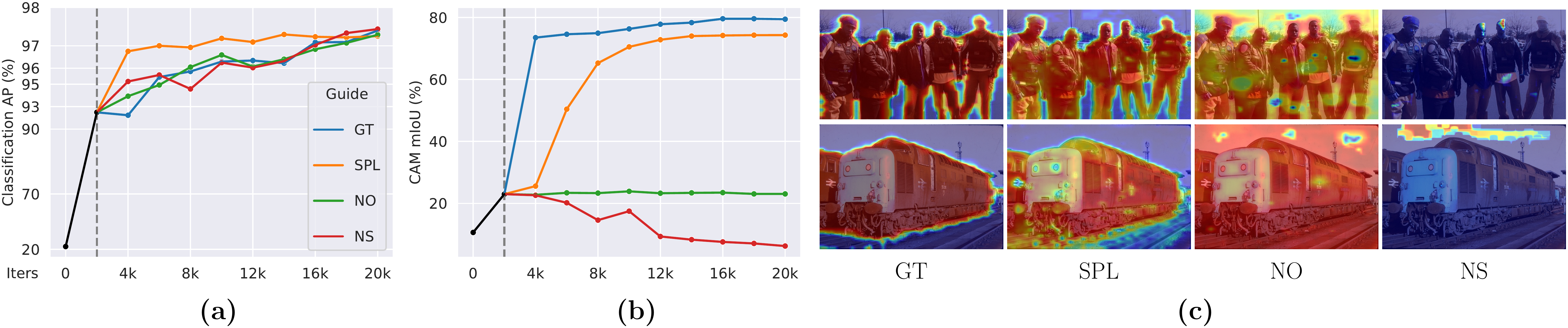}
  \caption{
    \capfont{
    \textbf{Oracle Experiments} on VOC. CAMs are guided by the ground truth (GT), proposed segmentation pseudo-labels (SPL), no guidance (NO) and random noise (NS).
\textbf{(a)}: classification performance; \textbf{(b)}: CAM quality;  \textbf{(c)} CAM visualization. All experiments employ 2k-iters warm-up before guidance is introduced.}
}
     \label{fig:oracle_exp}
\end{figure}

To demonstrate, we conduct oracle experiments wherein we supervise the output CAMs from a classifier with the ground truth segmentation (GT), enabling optimization of all elements in $F$.
For comparison, we conduct experiments where the CAMs are not guided (NO), and guided with {random noise masks (NS).} Results, shown in \cref{fig:oracle_exp}, demonstrate that different guidance for $M$ does not affect classification even for the NS group, as all experiment groups achieved over 97\% classification precision.
However, drastic differences can be observed \textit{w.r.t.} the quality of the CAMs. The GT group results in a notable quality improvement over the NO group, as shown in \cref{fig:oracle_exp}(b)(c).
In contrast, the NS group sabotages the CAMs. This suggests the stochasticity of CAMs and explains their variability and lack robustness, something also observed in \cite{ahn2018learning,wang2020self,chen2022c}.

Since relying on GT segmentation is not feasible in WSSS, we propose an alternative for guiding CAMs, employing predicted masks as segmentation pseudo-labels (SPL).
As shown in \cref{fig:oracle_exp}, a SPL-guided classifier yields CAMs that significantly outperform vanilla CAMs (NO), performing close to the oracle (GT).
Motivated by this, we introduce a co-training mechanism in which CAMs and predicted masks are optimized mutually without additional CAM refinement.

\subsection{Co-training with Swapping Assignments} \label{sec:cosa}

\begin{figure}[th]
\begin{center}
  \includegraphics[width=\linewidth]{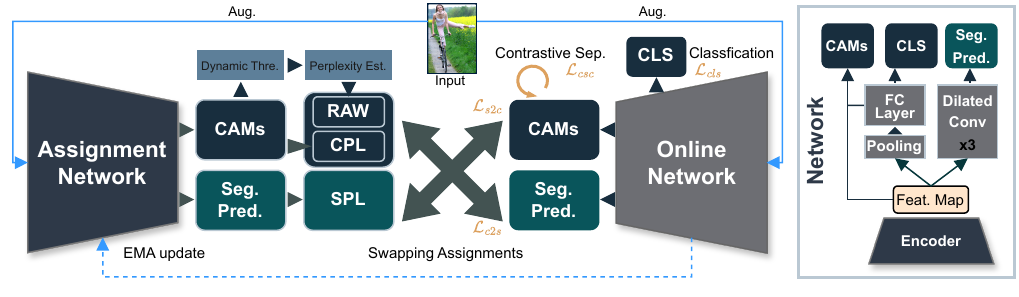}
\end{center}
\caption{
  \capfont
  {\textbf{Co-training with Swapping Assignments (CoSA)}.
  We propose an end-to-end dual-stream weakly-supervised segmentation framework{,} capable of co-optimizing the segmentation prediction and CAMs by leveraging the swapped assignments, namely CAM pseudo-labels (CPL) and segmentation pseudo-labels (SPL).
Our framework comprises two networks: {an assignment network (AN) and an online network (ON), where the AN is responsible for generating pseudo-labels for training the ON}. While the AN has identical architecture to the ON, it is updated through exponential moving average (EMA) of the ON. The diagram on the right provides an illustration of the architecture. Given weak-augmented images as input, the AN produces CPL to supervise segmentation in the ON{ ($\mathcal{L}_{\text{c2s}}$)}. During training, the CPL is softened by reliability-based adaptive weighting (RAW), formed based on {CAM perplexity estimation} and dynamic thresholding.
The AN also generates SPL which is utilized to supervise the CAMs{ ($\mathcal{L}_{\text{s2c}}$)}.
Further, the CAMs are regularized to contrastively separate the foreground from the background regions{ ($\mathcal{L}_{\text{csc}}$)}. Note that the ON is also trained for classification using the image-level class labels ($\mathcal{L}_{\text{cls}}$).
}}
\label{fig:overview}
\end{figure}

\noindent\textbf{Overall Framework.}
As shown in \cref{fig:overview}, CoSA contains two networks: an \textit{online} network (ON) and an \textit{assignment} network (AN). ON, parameterized by $\Theta$, comprises three parts: a backbone encoder, \texttt{FC} layers, and a segmentation head. AN has the same architecture as ON but uses different weights, denoted $\Theta^{\prime}$. ON is trained with the pseudo assignments generated by AN, while AN is updated by the exponential moving average of ON: $\Theta^\prime \leftarrow m\Theta^\prime+(1-m)\Theta,$ where $m \in [0,1]$ denotes a momentum coefficient. Consequently, the weights of AN represent a delayed and more stable version of the weights of ON, which helps to yield a consistent and stabilized learning target \cite{grill2020bootstrap}.

{A}n image and class label pair $(x, Y_{\text{gt}})$ is randomly sampled from a WSSS dataset $\mathcal{D}$.
CoSA utilizes two augmented views $\mathcal{T}_s(x)$ and $\mathcal{T}_w(x)$ as input for ON and AN, respectively, {representing} strong and weak image transformations. During training, AN produces CAMs $\mathcal{M}^\prime$ and segmentation predictions $\mathcal{S}^\prime$. The CAM pseudo-labels (CPL) and segmentation pseudo-labels (SPL) {are} generated by $\mathcal{M}^\prime$ and $\mathcal{S}^\prime$ after filtering with respect to $Y_{\text{gt}}$.
CPL and SPL are subsequently used as learning targets for supervising the segmentation predictions $\mathcal{S}$ and CAMs $\mathcal{M}$ from ON, respectively.

\vspace{3pt}\noindent\textbf{Swapping Assignments.}
Our objective is to co-optimize $\mathcal{S}$ and $\mathcal{M}$. {A} naive approach could enforce the learning objectives $\mathcal{S}\triangleq \mathcal{S}^\prime$ and $\mathcal{M}\triangleq \mathcal{M}^\prime$ as a knowledge distillation process \cite{hinton2015distilling}, where AN and ON play the roles of teacher and student. However, this assumes availability of a pretrained teacher which is not possible in WSSS settings.
Instead, we setup a swapped self-distillation objective:
\begin{equation} \label{eq:swap_obj}
  \small
  \mathcal{L}_{\text{swap}} = \mathcal{L}_{\text{c2s}}(\mathcal{S},\mathcal{M}^\prime) +\mathcal{L}_{\text{s2c}}(\mathcal{M},\mathcal{S}^\prime) ~,
\end{equation}
where $\mathcal{L}_{\text{c2s}}$ optimizes the segmentation performance given the CPL, and $\mathcal{L}_{\text{s2c}}$ assesses the CAM quality with respect to SPL.
Building on self-distillation \cite{caron2021emerging,oquab2023dinov2}, we present this swapped self-distillation framework tailored specifically to facilitate information exchange between the CAMs and segmentation.

\subsection{Segmentation Optimization}
\noindent\textbf{CAM2Seg Learning.}
Previous studies \cite{krahenbuhl2011efficient,ahn2018learning,araslanov2020single,ru2022learning} refine the CAMs to obtain pseudo-label, then perform pseudo-label to segmentation learning (PL2Seg).
As our guided-CAMs do not require extra refinement process, they can be directly employed as learning targets (CAM2Seg).
Nonetheless, CAMs primarily concentrate on the activated regions of the foreground while disregarding the background. As per the established convention \cite{wang2020self,cheng2023out,ru2023token}, a threshold value $\xi$ is employed for splitting the foreground and the background. Formally, our CAM pseudo-label (CPL) is given by:
\begin{equation}\label{eq:CPL}
  \small
    \hat{\mathcal{Y}}^{\text{CPL}}_{x,y}=\left\{
    \begin{aligned}
         & \mathtt{argmax}(\mathcal{M}_{x,y}^\prime)+1, & \text{if $\nu\geq\xi$,} \\
         & 0,                          & \text{if $\nu<\xi$,} \\
    \end{aligned}
    \right.,
\end{equation}
where $\nu\triangleq\mathtt{max}(\mathcal{M}_{x,y}^\prime)$ denotes the the maximum activation, $0$ denotes the background index.
Then, the CAM2seg learning objective $\mathcal{L}_{\text{c2s}}$ is cross entropy between ${\mathcal{Y}}^{\text{CPL}}$ and $\mathcal{S}$, {as with} the general supervised segmentation loss \cite{chen2017deeplab}.

\begin{figure}[t]
  \centering
  \includegraphics[width=1\linewidth]{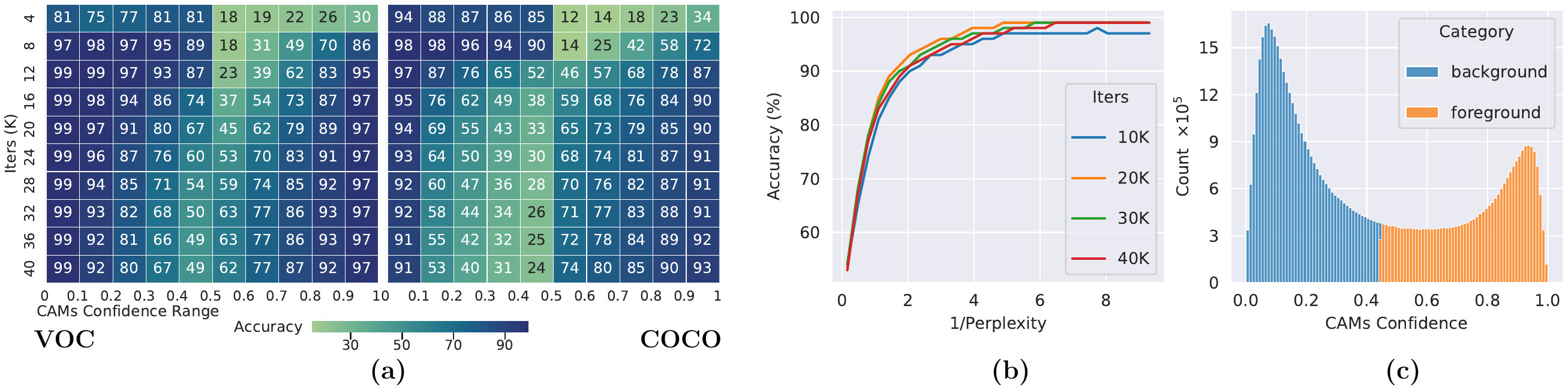}
  \caption{
    \capfont{
    {{\textbf{CPL Analysis}
  \textbf{(a)}: heatmap of CPL accuracy \textit{vs.} confident ranges (x-axis) for different time-steps (y-axis) for VOC and COCO.
  \textbf{(b)}: correlation between perplexity and accuracy of CPL for different time-steps.
  \textbf{(c)}: distribution of CAMs' confidence categorized by the proposed dynamic threshold on VOC. See \textit{Supp.} for COCO analysis.
}}}}
\label{fig:raw}
\end{figure}

\vspace{3pt}\noindent\textbf{Reliability based Adaptive Weighting (RAW).}
Segmentation performance depends heavily on the reliability of the pseudo-labels. Thus, it is important to assess their reliability. Existing methods use post-refinement to enhance pseudo-label credibility \cite{zhang2020reliability,araslanov2020single}. As CoSA can generate online CPL, we propose to leverage confidence information to compensate the CAM2Seg loss during training. Specifically, we propose to assess the perplexity scores for each pixel in $\hat{\mathcal{Y}}^{\text{CPL}}$ and leverage these scores to re-weight $\mathcal{L}_{\text{c2s}}$ for penalizing unreliable regions. However, estimating per-pixel perplexity is non-trivial.
Through empirical analysis, we observe a noteworthy association between the confidence values of CAMs and their accuracy at each time-step. This correlation suggests that regions with extremely low or high confidence exhibit higher accuracy throughout training, as shown in \cref{fig:raw}(a).
To quantitatively model perplexity, we make two assumptions:
i) the reliability of pseudo-labels is positively correlated with their accuracy,
and ii) the perplexity score is negatively correlated with the reliability.
Then, per-pixel perplexity of $\hat{\mathcal{Y}}^{\text{CPL}}_{x,y}$ is defined as:
\begin{equation}\label{eq:perplexity}
  \small
    \mathcal{P}_{x,y}=\left\{
    \begin{aligned}
         &\left[-\log\left(\lambda_{\alpha}(\nu-\xi)/(1-\xi)\right)\right]^{\lambda_\beta} & \text{if $\nu\geq\xi$,} \\
         &\left[-\log\left(\lambda_{\alpha}(\xi-\nu)/\xi\right)\right]^{\lambda_\beta}                          & \text{if $\nu<\xi$,} \\
    \end{aligned}
    \right.
\end{equation}
where the term within the logarithm denotes the normalized distance to $\xi$ in $[0,1]$. The logarithm ensures $\mathcal{P}_{x,y}\!\rightarrow\!+\infty$ as distance $\!\rightarrow\!0$, and $\mathcal{P}_{x,y}\!\rightarrow\!0$ as distance $\rightarrow \!1$. $\lambda_{\alpha}\in\mathbb{R}^{+}$ controls the perplexity score's minimum value and $\lambda_{\beta}\in\mathbb{R}^{+}$ determines the sharpness or smoothness of the distribution.
Higher $\mathcal{P}_{x,y}$ indicates confidence of $\hat{\mathcal{Y}}^{\text{CPL}}_{x,y}$ closer to threshold $\xi$. This observation is substantiated by \cref{fig:raw}(a), where confidence values near $\xi\!=\!0.5$ exhibit lower reliability. Furthermore, the correlation between perplexity and accuracy remains significant across various training time-steps and datasets, as depicted in \cref{fig:raw}(b).

Since we hypothesize negative reliability-perplexity correlation, the reliability score can be defined as the reciprocal of perplexity. To accommodate reliability variation for different input, we use the normalized reliability as the per-pixel weights for $\mathcal{L}_{\text{c2s}}$. This arrives our RAW-based CAM2Seg objective:
\begin{equation}\label{eq:c2s_pixel_loss_new}
  \small
  \mathcal{L}_{\text{c2s}}(x,y)\!\!=\!\!-
  \frac{|\mathcal{R}|}{\sum_{i,j \in \mathcal{R}} {(\mathcal{P}_{i,j}\mathcal{P}_{x,y})}^{-1}}
  \!\sum_{c=0}^{C}\!\Bigg[\!\mathbbm{1}\!\big[\hat{\mathcal{Y}}^{\text{CPL}}_{x,y}\!\!=\!\!c\big]\!\! \log\!\!\Bigg(\!\frac{\exp{\mathcal{S}^c_{x,y}}}{\sum_{k=0}^{C} \exp \mathcal{S}^k_{x,y}}\!\!\Bigg)\!\!\Bigg],
\end{equation}
where $|\mathcal{R}|$ represents total number of pixels in a batch.

\vspace{3pt}\noindent\textbf{Dynamic Threshold.}
Existing WSSS work \cite{ru2022learning,ru2023token} prescribes a fixed threshold to separate foreground and background, which neglects inherent variability due to prediction confidence fluctuation during training. Obviously, applying a fixed threshold in \cref{fig:raw}(a) is sub-optimal.

To alleviate this, we introduce dynamic thresholding. As shown in \cref{fig:raw}(c), the confidence distribution reveals discernible clusters. {W}e assume the foreground and background pixels satisfy a bimodal Gaussian Mixture distribution.
Then, the optimal dynamic threshold $\xi^{\star}$ is determined by maximizing the Gaussian Mixture likelihood:
\begin{equation} \label{eq:dynamic_thre}
  \small
  \begin{aligned}
  \xi^{\star}= \underset{\xi}{\mathtt{argmax}}
  \prod_{x \in \{\mathcal{M}^\prime \geq \xi\}}
  \tilde{\pi}_{fg} \mathcal{N}\left(x|\tilde{\mu}_{fg}, \tilde{\Sigma}_{fg}\right)
  + \prod_{x \in \{\mathcal{M}^\prime < \xi\}}
  \tilde{\pi}_{bg} \mathcal{N}\left(x|\tilde{\mu}_{bg}, \tilde{\Sigma}_{bg}\right) ~,
\end{aligned}
\end{equation}
where $\mathcal{N}(x|\mu,\Sigma)$ denotes the Gaussian function and $\pi$, $\mu$, $\Sigma$ {are} the weight, mean and covariance.
To avoid mini-batch bias, we maintain a queue to fit GMM, with the current $\mathcal{M}^{\prime}$ batch enqueued and the oldest dequeued. This facilitates establishment of a gradually evolving threshold, contributing to learning stabilization.

\subsection{CAM Optimization}
\begin{figure}[t]
  \centering
  \includegraphics[width=1\linewidth]{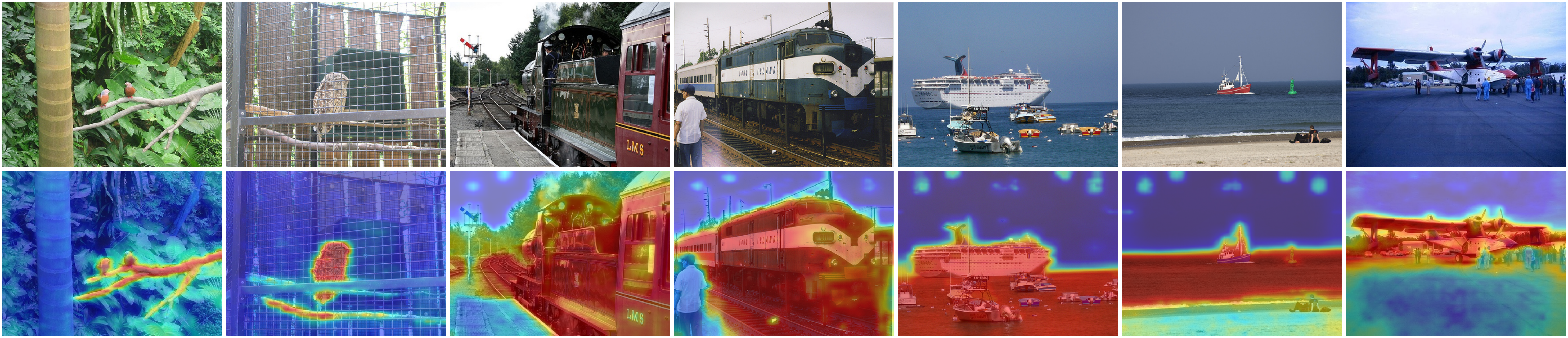}
  \caption{
    \capfont{
    {\textbf{Illustration of Coexistence Problem in CAMs.}
The first row shows the input images. The second row shows the coexistence problem e.g.  'bird' with 'branches', 'train' with 'railways' and 'boat' with 'the sea'.}}}
  \label{fig:coex}
\end{figure}

\noindent\textbf{Seg2CAM Learning.}
To generate SPL, segmentation predictions $\mathcal{S}^\prime$ are filtered by the weak labels $Y_{\text{gt}}$ and transformed into probabilities:
\begin{equation} \label{eq:SPL}
  \small
  \mathcal{S}^{\prime~c}_{x,y}=\left\{
    \begin{aligned}
    & -\infty,                       \!\!\!   &\text{if $Y_{\text{gt}}^{c}= 0 $,} \\
    & \mathcal{S}^{\prime~c}_{x,y}, \!\!\!& \text{if $Y_{\text{gt}}^{c}\neq 0 $,} \\
    \end{aligned}
    \right.
    \;\;\;\; \hat{\mathcal{Y}}^{\text{SPL}}_{x,y}=\mathtt{Softmax}(\frac{\mathcal{S}^{\prime}_{x,y}}{\tau}) ~,
\end{equation}
where $\tau$ denotes the temperature to sharpen $\hat{\mathcal{Y}}^{\text{SPL}}_{x,y}$. Let $\mathcal{R}$ be all the positions in SPL, then the Seg2CAM learning objective is defined as:
\begin{equation}
  \small
    \mathcal{L}_{\text{s2c}}= -\frac{1}{C|\mathcal{R}|}  \sum_{c=1}^{C} \sum_{x,y \in \mathcal{R}}\bigg[
      \hat{\mathcal{Y}}^{\text{SPL}}_{x,y}[c] \log(\sigma(\mathcal{M}^{c}_{x,y})) +  (1 - \hat{\mathcal{Y}}^{\text{SPL}}_{x,y}[c] ) \log(1-\sigma(\mathcal{M}^{c}_{x,y}))
\bigg] ~.
\end{equation}

\vspace{3pt}\noindent\textbf{Coexistence Problem in CAMs.}
{C}ertain class activations often exhibit large false positive regions, where objects are incorrectly merged with surroundings, as shown in \cref{fig:coex}. For instance, the classes `bird' and `tree branches', `train' and `railways', \textit{etc.} frequently appear together in VOC dataset. We refer to this issue as the coexistence problem. We hypothesize that the coexistence problem is attributed to three factors:
i) Objects that coexist in images, such as `tree branches', are not annotated \textit{w.r.t.} weak labels, which makes it challenging for a model to semantically distinguish coexistence.
ii) Training datasets lack sufficient samples for such classes.
iii) High-level feature maps, though rich in abstract representations and semantic information, lack essential low-level features such as edges, textures, and colors \cite{hariharan2015hypercolumns}.
Thus, CAMs generated from the last layer are poor in low-level information for segmenting objects. Conversely, segmenting objects with high-level semantics is hindered due to factors i) and ii).

\begin{figure}[t]
  \centering
  \includegraphics[width=0.75\linewidth]{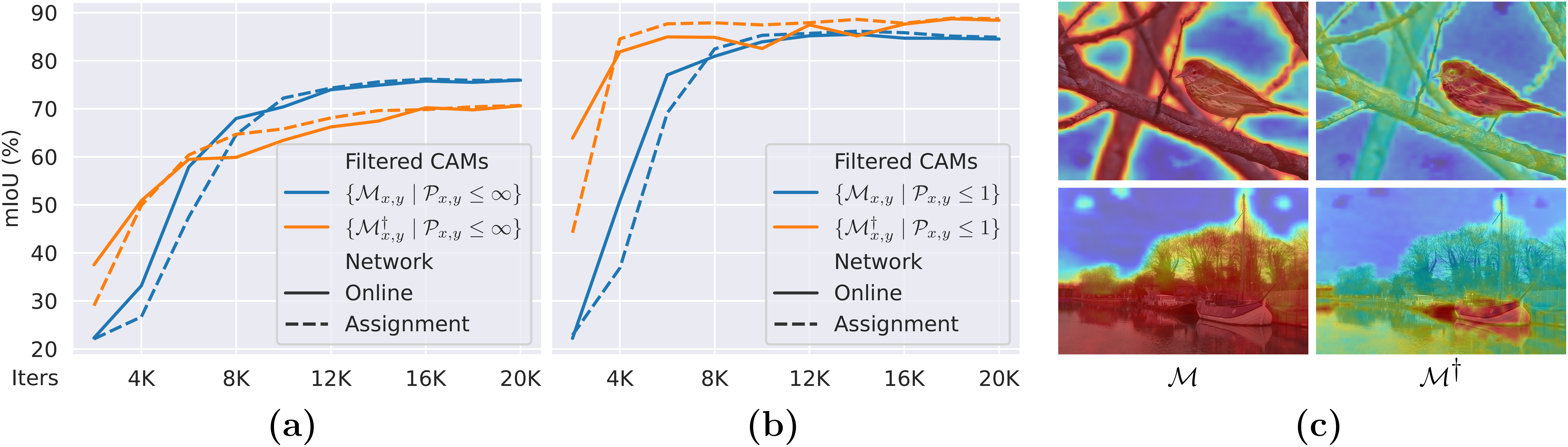}
  \caption{
    \capfont{
  {\textbf{$\mathcal{M}$ and $\mathcal{M}^\dagger$ Comparisons.} \textbf{(a)}: mIoU \textit{vs.} time-steps for $\mathcal{M}$ and $\mathcal{M}^\dagger$ on VOC \texttt{val}. \textbf{(b)}: same as \textbf{(a)} but filtered by perplexity. \textbf{(c)}: cases of coexistence in $\mathcal{M}$ but not in  $\mathcal{M}^\dagger$.}}}
  \label{fig:contrast}
\end{figure}

\vspace{3pt}\noindent\textbf{Contrastive Separation in CAMs.}
{W}e posit that the effective usage of low-level information can alleviate the coexistence problem.
Since shallower-layer feature is rich in low-level info \cite{zeiler2014visualizing}, we propose to extract CAMs $\mathcal{M}^\dagger$ from an earlier layer, and present its comparison with $\mathcal{M}$ in \cref{fig:contrast}, showing that directly substituting $\mathcal{M}$ with $\mathcal{M}^\dagger$ is not feasible due to the lower mIoU upperbound of $\mathcal{M}^\dagger$.
However, if we consider the confident regions in $\mathcal{M}$ and $\mathcal{M}^\dagger$, \textit{i.e.} filter by a low-pass perplexity, then $\{\mathcal{M}^\dagger_{x,y}\!\mid\!\mathcal{P}_{x,y}\!\leq\!\epsilon \}$ result in higher mIoU than $\{\mathcal{M}_{x,y}\!\mid\!\mathcal{P}_{x,y}\!\leq\!\epsilon \}$, as shown in \cref{fig:contrast}(b), where $\epsilon$ denotes a low-pass coefficient. Further,  we observe in some examples the presence of coexistence issues in $\mathcal{M}$ but absence in $\mathcal{M}^\dagger$ as shown in \cref{fig:contrast}(c). This suggests that $\mathcal{M}^\dagger$ performs worse than $\mathcal{M}$ in general, but better for those regions with low perplexity. Driven by these findings, we propose to regularize $\mathcal{M}$ by $\mathcal{M}^{ \dagger\prime}$ (from AN).
Specifically, $\mathcal{M}^{ \dagger\prime}$  after a low-pass filter are used to determine the positive  $\mathcal{R}^{+}_{i,j}$ and negative $\mathcal{R}^{-}_{i,j}$ regions:
\begin{equation} \label{eq:contrast_region}
  \small
  \begin{aligned}
    \mathcal{R}^{+}_{i,j}&= \Big\{(x,y) \mid \mathcal{P}_{x,y} \leq \epsilon, ~ \hat{y}^{\text{CPL}}_{x,y}=\hat{y}^{\text{CPL}}_{i,j}, (x,y)\neq (i,j) \Big\} \\
    \mathcal{R}^{-}_{i,j}&= \Big\{(x,y) \mid \mathcal{P}_{x,y} \leq \epsilon, ~ \hat{y}^{\text{CPL}}_{x,y}\neq\hat{y}^{\text{CPL}}_{i,j} \Big\}~,
  \end{aligned}
\end{equation}
where $(i,j)\!\in\!\Omega$, $\Omega\!=\!\{(x,y)\!\mid\!\mathcal{P}_{x,y}\!\leq\!\epsilon\}$ is low-perplexity region in $\mathcal{M}^{\dagger\prime}$, and $\hat{y}^{\text{CPL}}$ represents the CPL of $\mathcal{M}^{\dagger\prime}$.  Then, we have contrastive separation loss for $\mathcal{M}$:
\begin{equation} \label{eq:contrast}
  \small
  \mathcal{L}_{\text{csc}}=-\frac{1}{\lvert\Omega\rvert}\sum_{i,j \in \Omega}\frac{1}{\lvert \mathcal{R}^{+}_{i,j} \rvert}\sum_{x,y \in \mathcal{R}^{+}_{i,j}}
    \log \frac{L_{x,y}^{i,j}}
{L_{x,y}^{i,j} + K_{n,m}^{i,j}}
~,
\end{equation}
where $L_{x,y}^{i,j}\!=\!\exp( l_d(\mathcal{M}_{i,j},\mathcal{M}_{x,y})/\tau)$, $l_d(a,b)$ measures the similarity between (a,b), $\tau$ denotes the InfoNCE loss \cite{oord2018representation} temperature,  and
$K_{n,m}^{i,j}\!=\!\sum_{n,m\in\mathcal{R}^{-}_{i,j}}L_{n,m}^{i,j}$.

\vspace{3pt}\noindent\textbf{Overall Objectives.}
The objectives encompass the aforementioned losses and a further $\mathcal{L}_{\text{c2s}}^{\mathcal{M}^\dagger}$ to stabilize training and accelerate convergence, resulting in the CoSA objective:
\begin{equation}\label{eq:allloss}
  \small
  \mathcal{L}_{\text{CoSA}}\!\!=\! \mathcal{L}_{\text{cls}}\!+ \mathcal{L}_{\text{cls}}^{\mathcal{M}^\dagger}\!\! +\! \lambda_{\text{c2s}}\big(\mathcal{L}_{\text{c2s}}\!+\!\mathcal{L}_{\text{c2s}}^{\mathcal{M}^\dagger}\big)\!+\!\lambda_{\text{s2c}}\mathcal{L}_{\text{s2c}}+\!\lambda_{\text{csc}}\mathcal{L}_{\text{csc}}.
\end{equation}

\section{Experiments}
\subsection{Experiment Details and Results}
\noindent\textbf{Datasets.}
We evaluate on two benchmarks: {VOC} \cite{everingham2010pascal} and {COCO} \cite{lin2014microsoft}. VOC encompasses 20 categories with \texttt{train}, \texttt{val}, and \texttt{test} splits of 1464, 1449, and 1456 images. Following WSSS practice \cite{ahn2018learning,araslanov2020single,xu2022multi}, SBD \cite{hariharan2011semantic} is used to augment the \texttt{train} split to 10,582. COCO contains 80 categories with \texttt{train} and \texttt{val} splits of approx. 82K and 40K images. Our model is trained and evaluated using \emph{only} the image-level classification labels\footnote{Not available for VOC \texttt{test} split and so not used in evaluation.}, and employing mIoU as evaluation metrics.

\vspace{3pt}
\noindent\textbf{Implementation Details.}
Following \cite{ru2023token}, we use ImageNet pretrained ViT-base (ViT-B) \cite{dosovitskiy2021image} as the encoder. For classification, we use global max pooling (GMP) \cite{rossetti2022max} and the CAM approach \cite{zhou2016learning}.
For the segmentation decoder, we use LargeFOV \cite{chen2017deeplab}, as with \cite{ru2023token}. ON is trained with AdamW \cite{adamwoptim2017}. The learning rate is set to 6E-5 in tandem with polynomial decay. AN is updated with a momentum of $0.9994$. For preprocessing, the images are cropped to $448^2$, then weak/strong augmentations are applied (see \textit{Supp.}). The perplexity constants $(\lambda_{\alpha}, \lambda_{\beta})$  are set to $(0.8, 1)$, GMM-fitting queue length is $100$, and softmax temperature $\tau$ is $0.01$. The low perplexity threshold $\epsilon$  is set to $1$ and the loss weight factors $(\lambda_{\text{c2s}}, \lambda_{\text{s2c}}, \lambda_{\text{csc}})$ to $(0.1, 0.05, 0.1)$.

\vspace{3pt}
\noindent\textbf{Semantic Segmentation Comparison.}
We compare our method with existing SOTA WSSS methods on VOC and COCO for semantic segmentation in \cref{tab:semantic_seg}. CoSA achieves 76.2\% and 75.1\% on VOC12 \texttt{val} and \texttt{test}, respectively, surpassing the highest-performing single-stage model (ToCo) by 5.1\% and 2.9\%, as well as all multi-stage methods, including those with additional supervision. In the COCO evaluation, CoSA consistently outperforms other approaches, demonstrating a significant increase of 8.7\% over the top-performing single-stage methods. Further, there is a also 2.7\% improvement over the leading multi-stage method \cite{chen2023extracting}. While our primary goal is to provide an end-to-end WSSS solution, we also offer a multi-stage version of CoSA, denoted as \textbf{CoSA-MS} in \cref{tab:semantic_seg}, where various standalone segmentation networks are trained using our CPL. Our CoSA-MS models can also attains SOTA performance in multi-stage scenarios.

\begin{table}[h]
  \scriptsize
  \centering
  \setlength{\tabcolsep}{0.8mm}
  \begin{tabular}{lccccc}
    \toprule
    \multirow{2}{*}{\textbf{Methods} }                                                                        & \multirow{2}{*}{\textbf{\textit{Sup.}}} & \multirow{2}{*}{\textbf{\textit{Net.}}  } & \multicolumn{2}{c}{\textbf{VOC}} & {\textbf{COCO}}                                                                                                                                            \\ \cmidrule{4-5} \cmidrule{5-6}
                                                                                                              &                                         &                                           & \texttt{val}                     & \texttt{test}                                                                                                                              & \texttt{val}  \\ \midrule
    \multicolumn{4}{l}{\cellcolor[HTML]{ffffff}\textbf{\textit{Supervised Upperbounds}}.}                                                                                                                                                                                                                                                                                                           \\
    Deeplab \cite{chen2017deeplab}  \tiny TPAMI'2017                                                          & $\mathcal{F}$                           & R101                                      & 77.6                             & 79.7                                                                                                                                       & --            \\
    WideRes38 \cite{wu2019wider}     \tiny PR'2019                                                            & $\mathcal{F}$                           & WR38                                      & 80.8                             & 82.5                                                                                                                                       & --            \\
    ViT-Base  \cite{dosovitskiy2021image} \tiny ICLR'2021                                                     & $\mathcal{F}$                           & ViT-B                                     & 80.5                             & 81.0                                                                                                                                       & --            \\
    UperNet-Swin \cite{liu2021swin}     \tiny ICCV'2021                                                       & $\mathcal{F}$                           & SWIN                                      & 83.4                             & 83.7                                                                                                                                       & --            \\

    \midrule
    \multicolumn{4}{l}{\cellcolor[HTML]{ffffff}\textbf{\textit{Multi-stage Methods}}.}                                                                                                                                                                                                                                                                                                              \\
    L2G \cite{jiang2022l2g} \tiny CVPR'2022                                                                   & $\mathcal{I}+\mathcal{S}$               & R101                                      & 72.1                             & 71.7                                                                                                                                       & 44.2          \\
    Du \etal   \cite{du2022weakly}   \tiny CVPR'2022                                                          & $\mathcal{I}+\mathcal{S}$               & R101                                      & 72.6                             & 73.6                                                                                                                                       & --            \\
    CLIP-ES   \cite{lin2023clip}   \tiny CVPR'2023                                                            & $\mathcal{I}+\mathcal{L}$               & R101                                      & 73.8                             & 73.9                                                                                                                                       & 45.4          \\
    ESOL \cite{li2022expansion} \tiny NeurIPS'2022                                                            & $\mathcal{I}$                           & R101                                      & 69.9                             & 69.3                                                                                                                                       & 42.6          \\
    BECO \cite{rong2023boundary}  \tiny CVPR'2023                                                             & $\mathcal{I}$                           & R101                                      & 72.1                             & 71.8                                                                                                                                       & 45.1          \\
    Mat-Label   \cite{wang2023treating}   \tiny ICCV'2023                                                     & $\mathcal{I}$                           & R101                                      & 73.0                             & 72.7                                                                                                                                       & 45.6          \\
    \rowcolor[HTML]{eaeaea}
    \textbf{CoSA-MS}                                                                                          & $\mathcal{I}$                           & R101                                      & \textbf{76.5}                    & \textbf{75.3}\href{http://host.robots.ox.ac.uk:8080/anonymous/UEMZQP.html}{$^{\tiny{[1]}}$}                                                & \textbf{50.9} \\
    Xu \etal   \cite{xu2023learning}   \tiny CVPR'2023                                                        & $\mathcal{I}+\mathcal{L}$               & WR38                                      & 72.2                             & 72.2                                                                                                                                       & 45.9          \\
    W-OoD \cite{lee2022weakly} \tiny CVPR'2022                                                                & $\mathcal{I}$                           & WR38                                      & 70.7                             & 70.1                                                                                                                                       & --            \\
    MCT \cite{xu2022multi} \tiny CVPR'2022                                                                    & $\mathcal{I}$                           & WR38                                      & 71.9                             & 71.6                                                                                                                                       & 42.0          \\
    ex-ViT \cite{yu2023ex} \tiny PR'2023                                                                      & $\mathcal{I}$                           & WR38                                      & 71.2                             & 71.1                                                                                                                                       & 42.9          \\
    ACR-ViT \cite{kweon2023weakly} \tiny CVPR'2023                                                            & $\mathcal{I}$                           & WR38                                      & 72.4                             & 72.4                                                                                                                                       & --            \\
    MCT+OCR \cite{cheng2023out} \tiny CVPR'2023                                                               & $\mathcal{I}$                           & WR38                                      & 72.7                             & 72.0                                                                                                                                       & 42.0          \\
    \rowcolor[HTML]{eaeaea}
  \textbf{CoSA-MS}                                                                                            & $\mathcal{I}$                           & WR38                                      & \textbf{76.6}                    & \textbf{74.9}\href{http://host.robots.ox.ac.uk:8080/anonymous/BWWBSW.html}{$^{\tiny{[2]}}$}                                                & \textbf{50.1} \\
    ReCAM \cite{chen2022class} \tiny CVPR'2022                                                                & $\mathcal{I}$                           & SWIN                                      & 70.4                             & 71.7                                                                                                                                       & 47.9          \\
    LPCAM \cite{chen2023extracting} \tiny CVPR'2023                                                           & $\mathcal{I}$                           & SWIN                                      & 73.1                             & 73.4                                                                                                                                       & 48.3          \\
    \rowcolor[HTML]{eaeaea}
    \textbf{CoSA-MS}                                                                                          & $\mathcal{I}$                           & SWIN                                      & \textbf{81.4}                    & \textbf{78.4}\href{http://host.robots.ox.ac.uk:8080/anonymous/LGFR47.html}{$^{\tiny{[3]}}$}
                                                                                                              & \textbf{53.7}                                                                                                                                                                                                                                                                       \\
    \midrule
    \multicolumn{4}{l}{\cellcolor[HTML]{ffffff}\textbf{\textit{Single-stage (End-to-end) Methods}}.}                                                                                                                                                                                                                                                                                                \\
    RRM \cite{zhang2020reliability} \tiny AAAI'2020                                                           & $\mathcal{I}$                           & WR38                                      & 62.6                             & 62.9                                                                                                                                       & --            \\
    AFA \cite{ru2022learning} \tiny CVPR'2022                                                                 & $\mathcal{I}$                           & MiT-B1                                    & 66.0                             & 66.3                                                                                                                                       & 38.9          \\
    RRM \cite{zhang2020reliability}$^\dagger$ \tiny AAAI'2020                                                 & $\mathcal{I}$                           & ViT-B                                     & 63.1                             & 62.4                                                                                                                                       & --            \\
    ViT-PCM \cite{rossetti2022max} \tiny ECCV'2022                                                            & $\mathcal{I}$                           & ViT-B                                     & 69.3                             & --                                                                                                                                         & 45.0          \\
    ToCo \cite{ru2023token}  \tiny CVPR'2023                                                                  & $\mathcal{I}$                           & ViT-B                                     & 71.1                             & 72.2                                                                                                                                       & 42.3          \\
    SeCo \cite{yang2024separate}  \tiny CVPR'2024                                                             & $\mathcal{I}$                           & ViT-B                                     & 74.0                             & 73.8                                                                                                                                       & 46.7          \\
    \rowcolor[HTML]{eaeaea}
    \textbf{CoSA}                                                                                             & $\mathcal{I}$                           & ViT-B                                     & 76.2                             & 75.1\href{http://host.robots.ox.ac.uk:8080/anonymous/GOZOHI.html}{$^{\tiny{[4]}}$}                                                         & 51.0          \\
    \rowcolor[HTML]{eaeaea}
    \textbf{CoSA$^*$}                                                                                         & $\mathcal{I}$                           & ViT-B                                     & \textbf{76.4}                    & \textbf{75.2}\href{http://host.robots.ox.ac.uk:8080/anonymous/4SW3UJ.html}{$^{\tiny{[5]}}$}                                                & \textbf{51.1} \\
    \bottomrule
  \end{tabular}
  \caption{
    \captfont{
    \textbf{Weakly Supervised Semantic Segmentation Results}.
\textit{Sup.}: supervision type. \textit{Net.}: segmentation backbone. $\mathcal{F}$: Fully supervised, $\mathcal{I}$: Image-level labels, $\mathcal{S}$: Saliency maps, $\mathcal{L}$: language models. $*$ represents CRF \cite{chen2017deeplab} postprocessing results.}} \label{tab:semantic_seg}
\end{table}

\vspace{3pt}\noindent\textbf{CAM Quality Comparison.}
\cref{tab:CPL_eval} shows CoSA's CPL results compared with existing WSSS methods.
Our method yields 78.5\% and 76.4\% mIoU on \texttt{train} and \texttt{val}.
Notably, an ensemble of $\mathcal{M}^{\prime}$ and $\mathcal{M}^{\dagger \prime}$ improves performance to 78.9\% and 77.2\%, suggesting the activation of $\mathcal{M}^{\prime}$ is orthogonal to that of $\mathcal{M}^{\dagger \prime}$.

\begin{table}[hb]
  \addtolength{\tabcolsep}{2pt}
  \centering
  \center
\resizebox{0.9\columnwidth}{!}
  {
    \begin{tabular}{lccccccc}
      \toprule
      {\textbf{Method}} & ViT-PCM \cite{rossetti2022max} & ACR-ViT \cite{kweon2023weakly} & CLIP-ES     \cite{lin2023clip} & SeCo \cite{yang2024separate} & ToCo     \cite{ru2023token} & {CoSA} & {CoSA$^\bullet$}
      \\  \midrule
      \texttt{train}    & 71.4                           & 70.9                           & 75.0                           & 76.5                         & 73.6                        & 78.5   & \textbf{78.9}\\
      \texttt{val}      & 69.3                           & --                             & --                             & --                           & 72.3                        & 76.4   & \textbf{77.2}\\

      \bottomrule
\end{tabular}}
    \caption{
      \captfont{
      {\textbf{Comparisons of CPL}.
      All methods use ViT as the backbone for generating the CAMs on VOC dataset. $\bullet$ represents the ensemble of $\mathcal{M}^{\prime}$ and $\mathcal{M}^{\dagger \prime}$ in CoSA.
  }}
    \label{tab:CPL_eval}%
  }
\end{table}

\begin{figure}[t]
  \begin{minipage}[b]{0.7\textwidth}
  \centering
  \includegraphics[width=\linewidth]{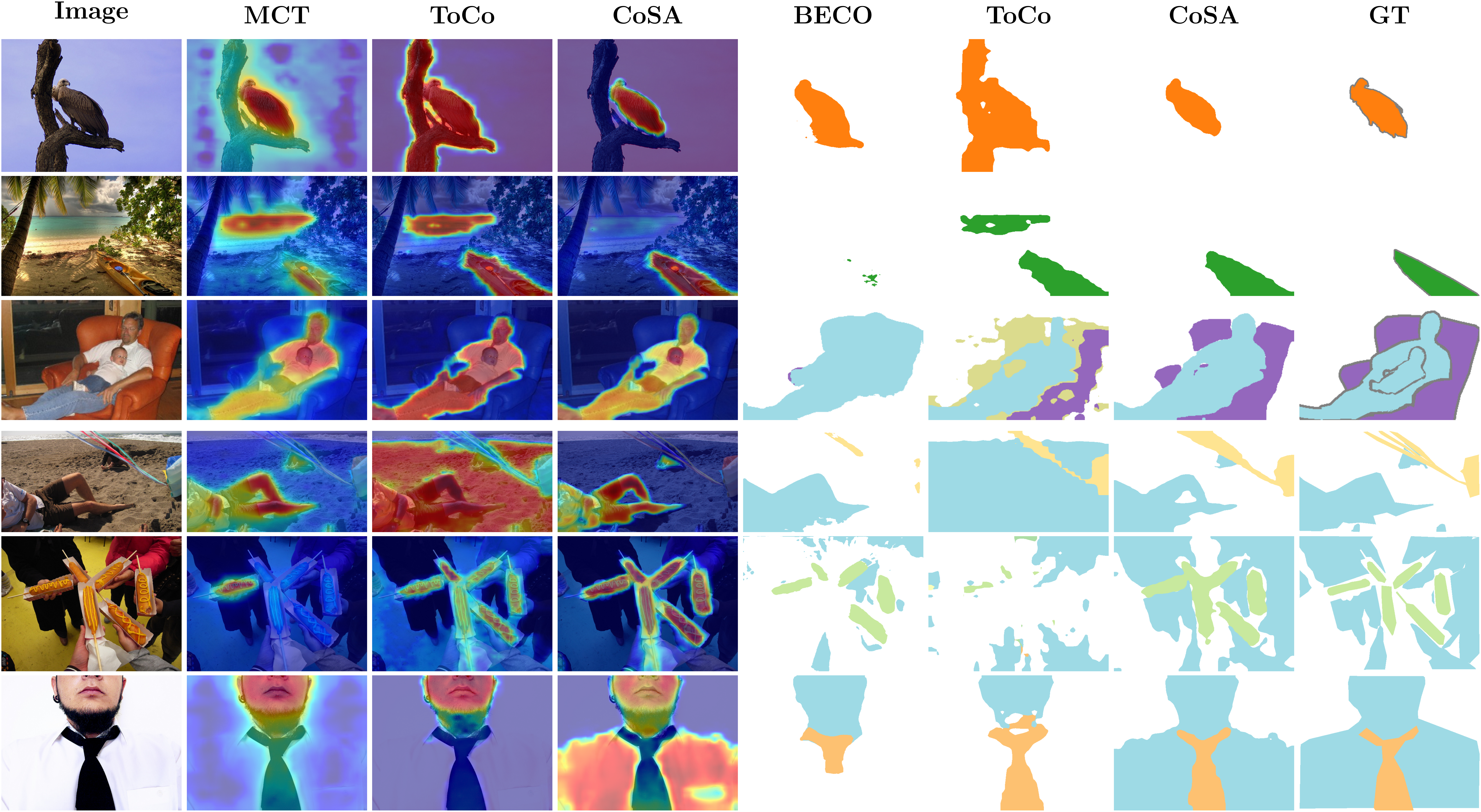}
  \caption{
    \capfont{
    {\textbf{Qualitative Comparison.}
The results are reported on the \texttt{val} splits of VOC (in \textit{R1} - \textit{R3}) and COCO (in \textit{R4} - \textit{R6}). The official codebases and provided weights for MCT \cite{xu2022multi}, BECO \cite{rong2023boundary}, and ToCo \cite{ru2023token} are used for this comparison.
(best viewed under zoom; see \textit{Supp.} for more).
}}}
\label{fig:comparison}
\end{minipage}\hfill
     \begin{minipage}[b]{0.28\textwidth}
        \centering
        \includegraphics[width=0.93\linewidth]{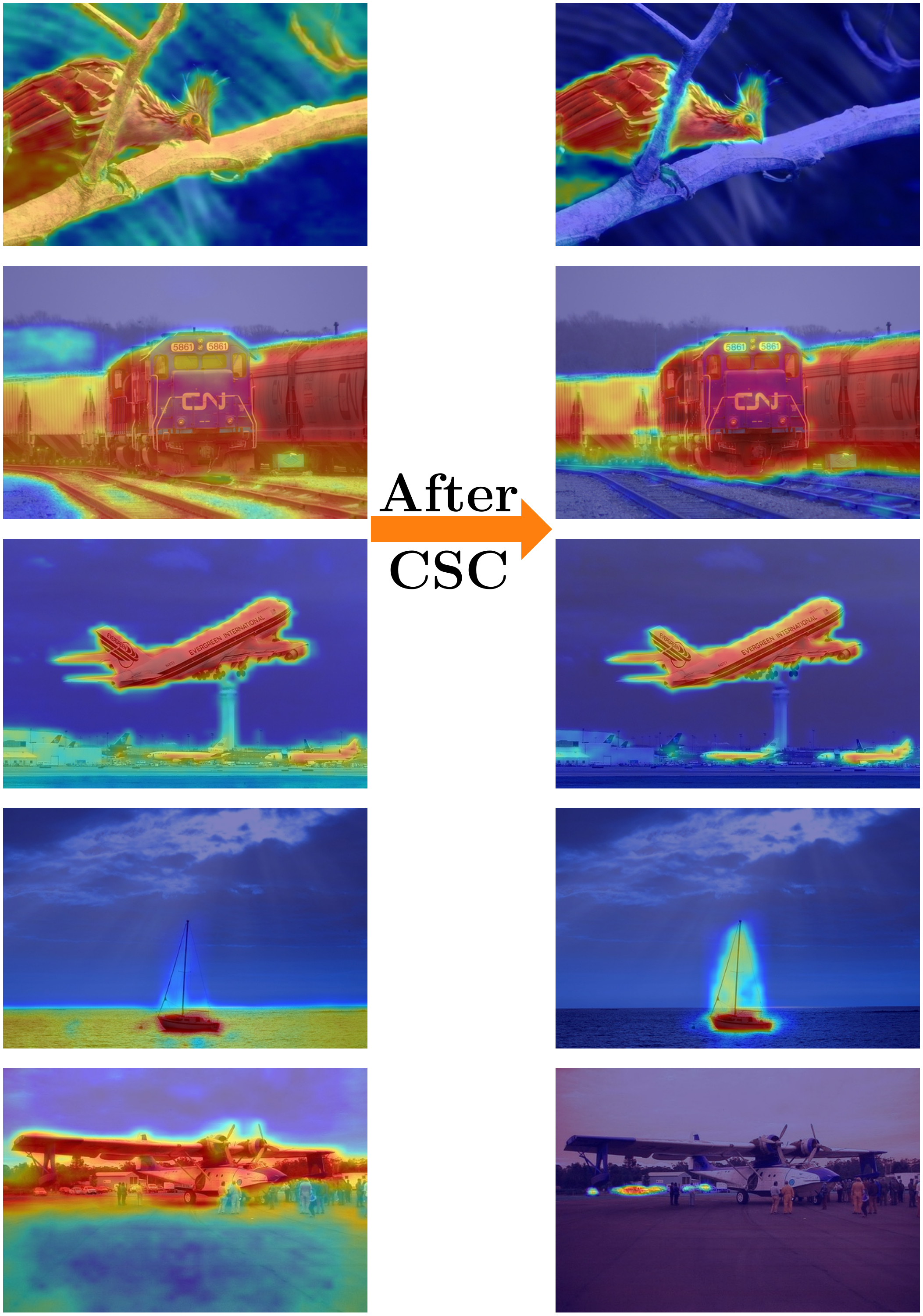}
        \caption{
    \capfont{
          {\textbf{Effect of CSC.}
      The class activation for bird, train, plane, boat and car are presented from top to bottom.
}}}
        \label{fig:AB_CSC}
     \end{minipage}
\end{figure}

\vspace{3pt}\noindent\textbf{Qualitative Comparison.}
\cref{fig:comparison} presents CAMs and segmentation visualizations, comparing with recent methods: MCT, BECO, and ToCo. As shown, our method can generate improved CAMs and produce well-aligned segmentation, exhibiting superior results in challenging segmentation problems with intra-class variation and occlusions.
In addition, CoSA performs well \textit{w.r.t.} the coexistence cases (\cref{fig:comparison} \textit{R1}, \textit{R2}), while existing methods struggle. Moreover, CoSA reveals limitations in the GT segmentation (\cref{fig:comparison} \textit{R4}).

  \begin{table}[th]
  \addtolength{\tabcolsep}{-0.5pt}
  \hspace{-5pt}
  \begin{minipage}[t]{0.53\textwidth}
    \begin{center}
      \scriptsize{\textbf{{(a)}}}\\
\resizebox{1\columnwidth}{!}{
    \begin{tabular}{ccccccccc}
        \toprule
                     &                                         &             &              &              &             &  & \multicolumn{2}{@{}c@{}}{\textbf{mIoU} \gain{\tiny{(inc.)}}} \\
    \cmidrule{8-9}
      \textbf{Base.} & \textbf{GC}                             & \textbf{SA} & \textbf{RAW} & \textbf{CSC} & \textbf{DT} &  & VOC                                                             & COCO\\
    \midrule
      \cmark         &                                         &             &              &              &             &  & 55.96~~~~~~~~~~~                                                & 37.32~~~~~~~~~~~~\\
      \cmark         & \cmark                                  &             &              &              &             &  & 63.09 \gain{\tiny{(+7.13)}}\hspace{3pt}
                     & 42.55 \gain{\tiny{(+5.23)}}~~~\\
      \cmark         & \cmark                                  & \cmark      &              &              &             &  & 64.41 \gain{\tiny{(+8.45)}}\hspace{3pt}

                     & 43.92 \gain{\tiny{(+6.60)}}~~~\\
      \cmark         & \cmark                                  & \cmark      & \cmark       &              &             &  & 68.22 \gain{\tiny{(+12.26)}}
                     & 45.39 \gain{\tiny{(+8.07)}}~~~\\
      \cmark         & \cmark                                  & \cmark      &              & \cmark       &             &  & 71.66 \gain{\tiny{(+15.70)}}
                     & 47.10 \gain{\tiny{(+9.78)}}~~~\\
      \cmark         & \cmark                                  & \cmark      & \cmark       & \cmark       &             &  & 75.54 \gain{\tiny{(+19.58)}}
                     & 49.67 \gain{\tiny{(+12.35)}}\\
      \cmark         & \cmark                                  & \cmark      & \cmark       & \cmark       & \cmark      &  & \textbf{76.19} \gain{\tiny{(+20.23)}}
                     & \textbf{51.00} \gain{\tiny{(+13.68)}}\\
        \bottomrule
                  \end{tabular}
                }
    \end{center}
\end{minipage}
  \begin{minipage}[t]{0.46\textwidth}
    \begin{center}
      \scriptsize{\textbf{{(b)}}}\\
\resizebox{1.05\columnwidth}{!}{
      \begin{tabular}{ccccccccc }
        \toprule
                       &             &             &              &              &             &  & \multicolumn{2}{@{}c@{}}{\textbf{mIoU} \dec{\tiny{(dec.)}}}   \\
    \cmidrule{8-9}
        \textbf{CoSA}  & \textbf{GC} & \textbf{SA} & \textbf{RAW} & \textbf{CSC} & \textbf{DT} &  & VOC                                                              & COCO\\
    \midrule
        \cmark         &             &             &              &              &             &  & \textbf{76.19}~~~~~~~                                            & \textbf{51.00}~~~~~~~ \\
        \cmark         &             &             &              &              & \xmark      &  & 75.54 \dec{\tiny{(-0.65)}}                                       & 49.67 \dec{\tiny{(-1.33)}}\\
        \cmark         &             &             &              & \xmark       &             &  & 69.89 \dec{\tiny{(-6.30)}}                                       & 45.95 \dec{\tiny{(-5.05)}}\\
    \cmark             &             &             & \xmark       &              &             &  & 72.45 \dec{\tiny{(-3.74)}}                                       & 47.83 \dec{\tiny{(-3.17)}}\\
    \cmark             &             & \xmark      &              &              &             &  & 72.10 \dec{\tiny{(-4.09)}}                                       & 49.04 \dec{\tiny{(-1.96)}}\\
    \cmark             & \xmark      &             &              &              &             &  & 74.12 \dec{\tiny{(-2.07)}}                                       & 49.67 \dec{\tiny{(-1.33)}}\\
        \bottomrule
                  \end{tabular}
                }
    \end{center}
\end{minipage}
  \caption{
    \captfont{
  {\textbf{Ablation Study on Contribution of Each Component.} \textbf{\textit{(a)}}: gradually add proposed components to baseline. \textbf{\textit{(b)}}: systematically exclude components from CoSA. \textbf{GC}: Guided CAMs, \textbf{SA}: Swapping Assignments, \textbf{RAW}: Reliability based Adaptive Weighting, \textbf{CSC}: Contrastive Separation in CAMs, and \textbf{DT}: Dynamic Threshold. \textbf{mIoU} is reported on PASCAL VOC12 and COCO \texttt{val} splits. }}}
\label{tab:module}
\end{table}

\subsection{Ablation Studies}

\noindent\textbf{CoSA Module Analysis.}
We begin by employing CAMs directly as the supervision signal for segmentation, akin to \cite{zhang2020reliability}, albeit without refinement, and gradually apply CoSA modules to this baseline. As shown in \cref{tab:module}{(a)}, the mIoUs progressively improve with addition of our components. Further, we examine the efficacy of each CoSA component. As shown in \cref{tab:module}{(b)}, the elimination of each component results in deteriorated performance, most notably for CSC.
\begin{table}[ht]
  \begin{minipage}[t]{0.30\linewidth}
  \centering
  \scriptsize{\textbf{(a)}}\\
  \resizebox{1.05\columnwidth}{!}{
  \begin{tabular}{@{} cc c cc@{}}
    \toprule
    \textbf{Source}  & \textbf{Detach} &  & \texttt{train}                             & \texttt{val}  \\
    \midrule

     GT              & None            &  & 83.99                                      & 80.16 \\
    \midrule

     NO              & --              &  & 72.28                                      & 71.38 \\
     SPL             & $F$             &  & 74.19                                      & 73.36 \\
     SPL             & $W_{\text{fc}}$ &  & 78.05                                      & 76.15  \\
     SPL             & None            &  & \textbf{78.54}                             & \textbf{76.37}  \\
    \bottomrule
  \end{tabular}
}
\end{minipage}
\hfill
\begin{minipage}[t]{0.30\linewidth}
  \centering
  \scriptsize{\textbf{(b)}}\\
  \resizebox{0.95\columnwidth}{!}{
  \begin{tabular}{@{} c c cc@{}}
    \toprule
    \textbf{Method}               &  & \textbf{C-mIoU} & \textbf{mIoU}                                 \\
    \midrule
    FPR  \cite{chen2023fpr}~~     &  & 53.09           & 53.34                                         \\
    ToCo  \cite{ru2023token}      &  & 63.62           & 72.33                                         \\
    SeCo  \cite{yang2024separate} &  & 73.18           & 73.63                                         \\
    \midrule
    w/o CSC                       &  & 62.61           & 67.82                                         \\
    w/ CSC                        &  & \textbf{82.34}  & \textbf{76.37}   \\
    \bottomrule
  \end{tabular}
}
\end{minipage}
\begin{minipage}[t]{0.38\linewidth}
 \addtolength{\tabcolsep}{-0pt}
 \centering
      \scriptsize{\textbf{(c)}}\\
  \resizebox{1.\columnwidth}{!}{
        \begin{tabular}{l c c | c c }
    \toprule
                          & \multicolumn{2}{@{}c@{}|}{\textbf{mIoU}\tiny{(\%)}} & \multicolumn{2}{@{}c@{}}{\textbf{Speed}} \\
        \textbf{Use CRF?} & \cmark                                              & \xmark                                      & \cmark & \xmark               \\
        \midrule
        BECO-R101         & 72.1                                                & 70.9\tiny{\dec{{(-1.2)}}}                   & 1.95   & 4.94\\
        COSA-R101         & 76.5                                                & 76.4\tiny{\dec{{{(\textbf{-0.1})}}}}        & 2.36   & \textbf{9.60 }\\
        ToCo-ViT          & 71.1                                                & 69.2\tiny{\dec{{(-1.9)}}}                   & 1.82   & 3.99 \\
        COSA-ViT          & 76.4                                                & 76.2\tiny{\dec{{{(\textbf{-0.2})}}}}        & 1.83   & \textbf{4.11}\\
    \bottomrule
        \end{tabular}
      }
\end{minipage}
  \caption{{
    \captfont{
      \textbf{Ablation Study of GC, CSC and CRF.}
      \textbf{(a)}: CPL performance comparison on VOC.
      \textbf{Source}: guidance sources. \textbf{Detach}: stop gradient in GC for feature map $F$ or $W_{\text{fc}}$.
      \textbf{(b)}: CPL performance comparison.
      FPR, ToCo and SeCo results are based on their code and weights.
      \textbf{C-mIoU}: mIoU for classes with coexistence.
\textbf{(c)}: CRF Impact. Best speed-accuracy tradeoff is achieved without using CRF. Inference speeds (FPS) are tested on RTX 3090.
}}}
  \label{tab:ab_gc_csc}
\end{table}

\begin{figure}
     \centering
     \begin{minipage}[t]{0.48\textwidth}
        \centering
        \includegraphics[width=\linewidth]{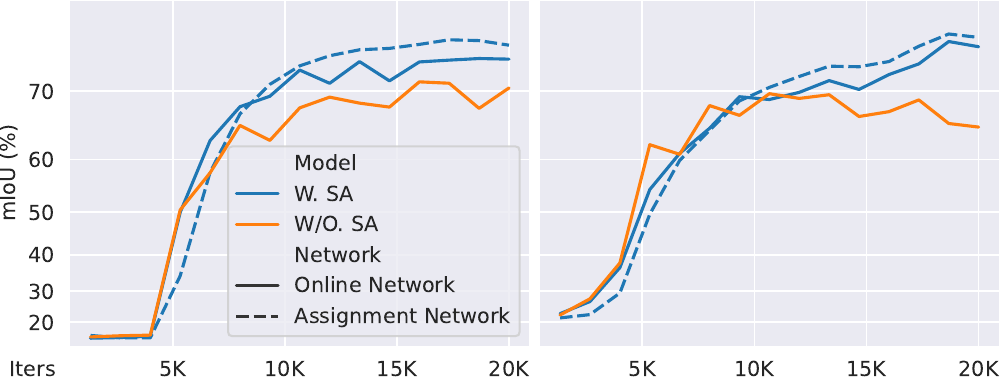}
        \caption{
    \capfont{
          {\textbf{Ablative Study of SA.} The performance of SPL (\textit{left}) and CPL (\textit{right}) \textit{w.r.t.} iterations on VOC \texttt{val} set are shown for CoSA with or without SA.
      }}}\label{fig:AB_SA}
     \end{minipage}
     \hfill
     \begin{minipage}[t]{0.48\textwidth}
          \centering
          \includegraphics[width=1\linewidth]{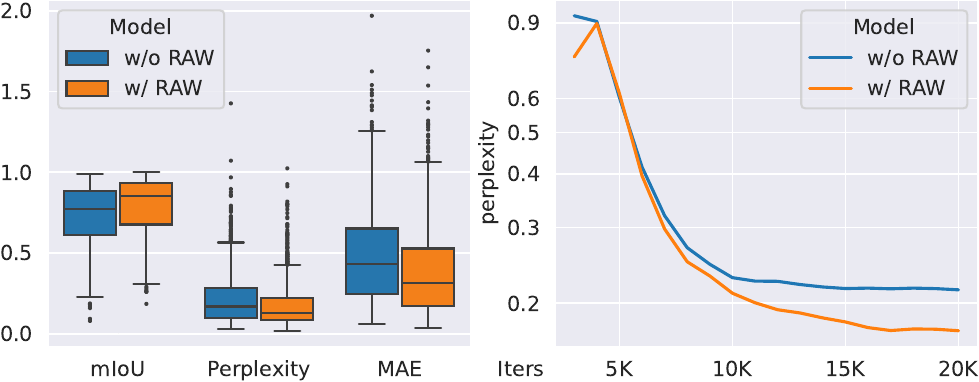}
          \caption{
    \capfont{
            {{\textbf{Ablation Study of RAW.}
              (\textit{left}) boxplot of mIoU, perplexity and MAE to (1,0) for individual CPLs on VOC \texttt{val}.
      (\textit{right}) perplexity reduction over times.}}}}
        \label{fig:AB_RAW}
     \end{minipage}
\end{figure}

\vspace{3pt}\noindent\textbf{Impact of Guided CAMs.}
Our model is compared with a baseline \cite{zhang2020reliability} that directly uses CAMs as CPL. As shown in \cref{tab:ab_gc_csc}(a), our guided CAMs notably enhance CPL quality by 6.26\% and 4.99\% for \texttt{train} and \texttt{val} splits. {Further, we conduct experiments to ascertain the extent to which the two CAM components, feature $F$ and weights $W_{\text{fc}}$, exert greater impact on guiding CAMs.} As shown, the deteachment of $F$ results in 74.19\% and 73.36\%, but $W_{\text{fc}}$ can decrease the results slightly to 78.05\% and 76.37\%.
This suggests that guiding CAMs primarily optimizes feature maps, verifying our hypothesis of the non-deterministic feature map contributing to the stochasticity of CAMs in \cref{sec:gcam}.

\vspace{3pt}\noindent\textbf{Impact of Swapping Assignments (SA).}
\cref{tab:module}(b) suggests that eliminating SA results in significant mIoU decreases, highlighting the importance of this training strategy.
Further examination of the ON and AN \textit{w.r.t.} SPL and CPL indicates that, in later training stages, AN consistently outperforms ON for both SPL and CPL, as shown in \cref{fig:AB_SA}, due to AN performing a form of model ensembling similar to Polyak-Ruppert averaging \cite{polyak1992acceleration,ruppert1988efficient}.
We observe a noticeable disparity of mIoUs between two ONs (solid orange line \textit{vs.} solid blue line in \cref{fig:AB_SA}), which may be attributed to the superior quality of CPL and SPL from the AN facilitating a more robust ON for CoSA. {The momentum framework, originally introduced to mitigate noise and fluctuations of the online learning target \cite{grill2020bootstrap,caron2021emerging}, is used for info exchange across CAMs and segmentation in CoSA.
}

\vspace{3pt}\noindent\textbf{Impact of RAW.}
\cref{tab:module}(b) shows notable mIoU reduction without RAW.
We conduct further studies to investigate its effect on perplexity reduction.
The boxplot in \cref{fig:AB_RAW} suggests that RAW leads to higher mIoU but lower perplexity.
\cref{fig:AB_RAW}(\textit{right}){ illustrates a faster decrease in perplexity} when RAW is used, affirming its impact on perplexity reduction.

\vspace{3pt}\noindent\textbf{Impact of CSC.}
Our CSC is introduced to address the coexistence issues. We establish C-mIoU to measure the CAM quality for those coexistence-affected classes. As shown in \cref{tab:ab_gc_csc}(b), applying CSC sees a boost in C-mIoU and mIoU, which surpass the existing methods. Some visual comparisons are given in \cref{fig:AB_CSC}.

\vspace{3pt}\noindent\textbf{Impact of Dynamic Threshold.}
We evaluate CoSA using some predetermined thresholds, comparing them with one employing dynamic threshold on VOC \texttt{val} split (see \textit{Supp.} for results).
The performance is sensitive to the threshold,
but dynamic thresholding achieves 0.65\% increased performance over the best manual finetuning while saving 80\% of hyper-parameter searching time.

\subsection{Further Analysis}
\noindent\textbf{Training and Inference Efficiency Analysis.}
Unlike multi-stage approaches, CoSA can be trained end-to-end efficiently.
Compared to BECO \cite{rong2023boundary}, our method is 240\% faster in training, uses 50\% fewer parameters, and yields a 4.3\% higher mIoU on VOC \texttt{test}.  Please refer to \textit{Supp.} for more discussion.
At inference time, we find that CRF post-processing, which is commonly adopted for refining masks \cite{cheng2023out,rong2023boundary} or the CAMs \cite{rossetti2022max,xu2022multi,zhang2020reliability}, can greatly slow down the inference speed.
As our aim is to develop a fully end-to-end WSSS solution, incorporating CRF post-processing contradicts this principal.
Through our experiments, we show that CoSA does not heavily depend on CRF: incorporating CRF results in marginal improvement of $0.2\%$, $0.1\%$, and $0.1\%$ for VOC \texttt{val}, VOC \texttt{test}, and COCO \texttt{val}, respectively (\cref{tab:semantic_seg}). Conversely, eliminating CRF in CoSA can greatly speed up inference (a noteworthy 307\% and 165\% $\uparrow$) and achieve better speed-accuracy tradeoff as suggested in \cref{tab:ab_gc_csc}(c).

\vspace{3pt}\noindent\textbf{Hyper-parameter Sensitivity.}
We apply grid search strategy to explore the hyper-parameters in CoSA. The analysis of parameters, such as \textit{perplexity filter}, \textit{loss weights}, \textit{momentum} are discussed in \textit{Supp.} CoSA maintains consistent with variations of those parameters, which demonstrate its robustness.

\section{Conclusion}
This paper presents an end-to-end WSSS method: Co-training with Swapping Assignments (CoSA), which eliminates the need for CAM refinement and enables concurrent CAM and segmentation optimization.
Our empirical study reveals the non-deterministic behaviors of CAMs and that proper guidance can mitigate such stochasticity, leading to substantial quality enhancement.
We propose explicit CAM optimization leveraging segmentation pseudo-labels in our approach, where a dual-stream model comprising an online network for predicting CAMs and segmentation masks, and an ancillary assignment network providing swapped assignments (SPL and CPL) for training, is introduced. We further propose three techniques within this framework: RAW, designed to mitigate the issue of unreliable pseudo-labels; contrastive separation, aimed at resolving coexistence problems; and a dynamic threshold search algorithm.
Incorporating these techniques, CoSA outperforms all SOTA methods on both VOC and COCO WSSS benchmarks while achieving exceptional speed-accuracy trade-off.

\section*{Acknowledgments}
The work is supported by European Research Council under the European Union’s Horizon 2020 research and innovation programme (GA No 787768).

\bibliographystyle{splncs04}
\bibliography{main}

\appendix

\section{Further Analysis}

\begin{figure}[ht]
  \centering
        \includegraphics[width=0.62\linewidth]{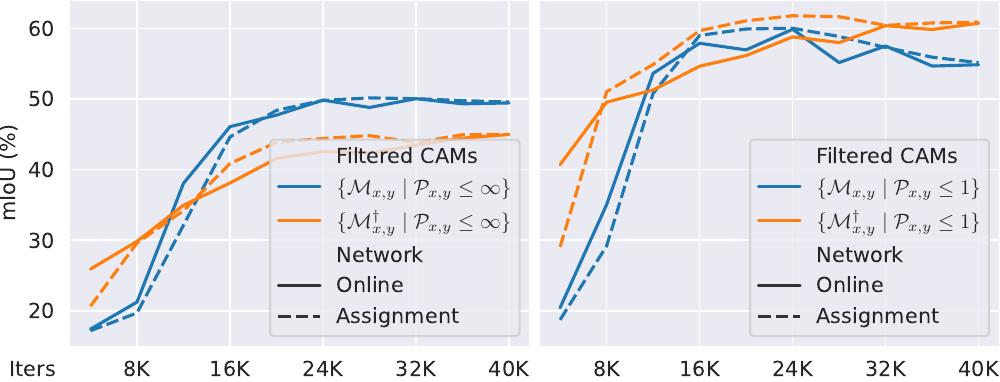}
        \caption{\textbf{$\mathcal{M}$ and $\mathcal{M}^{\dagger}$ Comparisons on COCO.}
          (\textit{left}): mIoU vs. time-steps for $\mathcal{M}$ and $\mathcal{M}^{\dagger}$ on COCO
\texttt{val}. (\textit{right}): same as (\textit{left}) but filtered by perplexity.
      }
    \label{fig:coco_cam}
\end{figure}

\vspace{5pt}\noindent \textbf{Contrastive Separation Analysis.}
\cref{fig:coco_cam} shows the analysis of contrastive separation on COCO. A similar trend between $\mathcal{M}$ and $\mathcal{M}^{\dagger}$ is observed as on VOC.
This suggests that the distinct relationship between $\mathcal{M}$ and $\mathcal{M}^{\dagger}$ extends beyond the VOC dataset to encompass a broader dataset as well.

\begin{figure}[ht]
  \centering
        \includegraphics[width=0.65\linewidth]{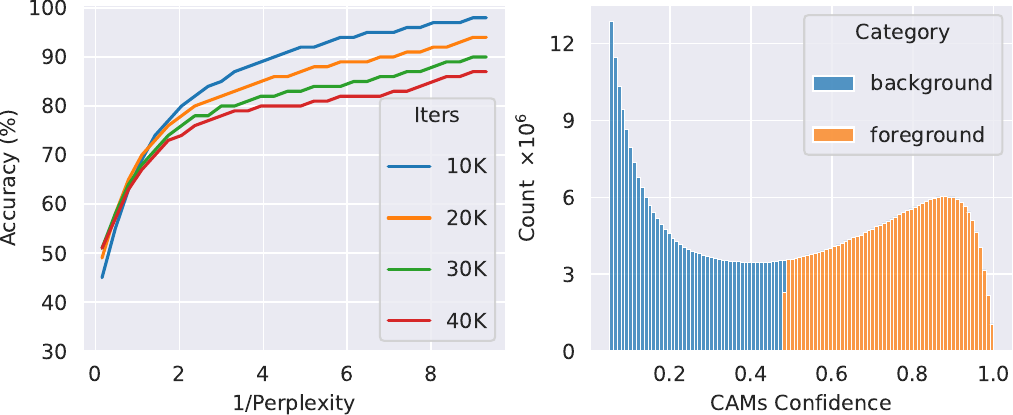}
        \caption{\textbf{CPL Analysis on COCO}
      (\textit{left}) correlation between perplexity and accuracy of CPL for different time-steps. (\textit{right}) distribution of CAMs’ confidence
categorized by the proposed dynamic threshold
      }
        \label{fig:coco_cpl}
\end{figure}

\vspace{5pt}\noindent\textbf{CPL Analysis.}
While the proposed dynamic learning techniques in Sec. 3.3, namely reliability based adaptive weighting and dynamic threshold, were originally inspired by the VOC dataset, we demonstrate their applicability on COCO here. As shown in \cref{fig:coco_cpl}(\textit{left}), the negative correlation between perplexity and accuracy remains significant across various training time-steps,  highlighting our purposed perplexity estimation method can be used to penalize inaccurate regions in our CAM2Seg loss. \cref{fig:coco_cpl}(\textit{right}) also suggests that the confidence distribution on the COCO dataset shows discernible clusters.

\vspace{5pt}\noindent\textbf{Impact of CRF.}
The conditional random field (CRF) proposes to optimize the segmentation by utilizing the low-level information obtained from the local interactions of pixels and edges \cite{chen2017deeplab}.
Traditionally, a manually designed CRF postprocessing step has been widely adopted for refining segmentation \cite{cheng2023out,rong2023boundary} or CAMs \cite{rossetti2022max,xu2022multi,zhang2020reliability}.
As our aim is to develop a fully end-to-end WSSS solution, incorporating CRF postprocessing contradicts this principal. Through our experiments, we demonstrate that CoSA, unlike other single-stage methods, does not heavily depend on CRF. Our results indicate that incorporating CRF results in marginal improvement of $0.2\%$, $0.1\%$, and $0.1\%$ for VOC \texttt{val}, VOC \texttt{test}, and COCO \texttt{val}, respectively, as presented in Tab. 2 of the main paper.
\cref{tab:ab_crf}(a) suggests that in comparison to other SOTA models, our CoSA exhibits a lesser dependency on the CRF postprocessing.
On the contrary, eliminating the CRF step leads to a noteworthy enhancement of 165\% in terms of inference speed, as demonstrated in \cref{tab:ab_crf}(b).

  \begin{table}[ht]
  \begin{minipage}[t]{0.6\linewidth}
    \begin{center}
      \textbf{{(a)}}\\
    \begin{tabular}{lcc}
        \toprule
         \textbf{Method}                                                       & \textbf{w/o  CRF} & \textbf{w/ CRF}              \\
        \midrule
         AFA \cite{ru2022learning}& 63.8 & 66.0 \scriptsize{\gain{(+2.2)}} \\
         VIT-PCM \cite{rossetti2022max}& 67.7 & 71.4 \scriptsize{\gain{(+3.7)}} \\
         ToCo \cite{ru2023token}& 69.2& 71.1 \scriptsize{\gain{(+0.9)}} \\
         SeCo \cite{yang2024separate}& 72.2& 73.7 \scriptsize{\gain{(+1.5)}} \\
      CoSA & 76.2 & 76.4 \scriptsize{\gain{(+0.2)}} \\
        \bottomrule
                  \end{tabular}
    \end{center}
\end{minipage}
  \begin{minipage}[t]{0.38\linewidth}
    \begin{center}
      \textbf{{(b)}}\\
      \begin{tabular}{l c }
        \toprule
                 CoSA & \textbf{Speed}              \\
        \midrule
        w/o CRF & 4.11 imgs/s \\
        w/ CRF & 1.83 imgs/s\\
        \bottomrule
                  \end{tabular}
    \end{center}
\end{minipage}
\vspace{5pt}
\caption{\textbf{CRF Impact.}
\textbf{{(a)}}: Comparisons of CRF impact on SOTA single-stage WSSS methods on VOC \texttt{val}. \textbf{{(b)}}: Inference speed with and without CRF. Speed tested using a RTX 3090.
}
\label{tab:ab_crf}
\end{table}

\vspace{5pt}\noindent\textbf{Training Efficiency Study.}
Unlike multi-stage approaches, CoSA is extremely efficient in training.
It can be trained end-to-end efficiently. When training a semantic segmentation model with weak labels on the VOC dataset, our method requires a mere 8.7 hours of training time and a total of 92M parameters.
In contrast, MCT \cite{xu2022multi} would necessitate approximately 231\% more time (20.1hrs $\uparrow$) and 173\% more parameters (159M $\uparrow$) for the same task, and BECO \cite{rong2023boundary} would require around 240\% more time (20.9hrs $\uparrow$) and 50\% more parameters (46M $\uparrow$). When compared to the single-stage method, CoSA also demonstrate its advantage in speed-accuracy trade-off. Further details regarding the efficiency study can be found in \cref{tab:ab_speed}.

\begin{table}[ht]
        \centering

\resizebox{1\columnwidth}{!}{
        \begin{tabular}{lcccccc}
        \toprule
        \multirow{2}{*}{}   & CAMs   & CAMs  & Seg. && \multirow{2}{*}{Total}  & \multirow{2}{*}{mIoU}\\
         & Generation & Refinement & Training\\
         \cmidrule{2-4}
        MCT\cite{xu2022multi} & 2.2hrs \gray{(21M)}  &  11.1hrs \gray{(106M)}   & 15.5hrs \gray{(124M)} && 28.8hrs \gray{(251M)} & 71.6\\
        BECO\cite{rong2023boundary} & 0.9hrs \gray{(23M)} & 6.5hrs ~\gray{(24M)} & 22.2hrs ~\gray{(91M)} && 29.6hrs \gray{(138M)} & 71.8\\
        \midrule
        ToCo\cite{ru2023token} & \multicolumn{3}{l}{~~~~~~~~~~~~~~~~~~~~~~~~~~~~9.9hrs~ \gray{(98M)}~~~~~~~~~~~~~~~~~~~~~~~~~~~~} && 9.9hrs~ \gray{(98M)} & 72.2 \\
        SeCo\cite{yang2024separate} & \multicolumn{3}{l}{~~~~~~~~~~~~~~~~~~~~~~~~~~~~8.8hrs~ \gray{(98M)}~~~~~~~~~~~~~~~~~~~~~~~~~~~~} && 8.8hrs~ \gray{(98M)} & 73.8 \\
        \midrule
        &  \multicolumn{3}{c}{CAMs and Seg. Co-optimization} &  \\
         \cmidrule{2-4}
        CoSA & \multicolumn{3}{c}{8.7hrs~ \gray{(92M)}} && \textbf{8.7hrs \gray{(92M)}} & \textbf{75.1} \\
        \bottomrule
        \end{tabular}
    }
    \caption{\textbf{Training Speed and Parameters Comparisons.}
    We report the detailed training time, parameters and final mIoU on VOC \texttt{test} split for MCT, BECO, ToCo and our CoSA.
  All methods are tested using the same machine with a single 3090 GPU.
  The official MCT, BECO and ToCo code repositories are utilized in this study.
}
    \label{tab:ab_speed}
\end{table}

\begin{algorithm}[ht]
  \scriptsize
\caption{CoSA Training Pseudo Code}\label{alg}
\begin{algorithmic}[1]
\State \textbf{Require:} {$\mathcal{D}$} \Comment{ image-level classification dataset}
\State \textbf{Require:} {$\mathcal{F}_\Theta,~\mathcal{F}_{\Theta^{\prime}}$} \Comment{ online network parameterized by $\Theta$ and assignment network by $\Theta^{\prime}$}
\State {$\mathcal{F}_\Theta$} $\gets$ \texttt{Init},~~~{$\mathcal{F}_{\Theta^{\prime}}$} $\gets$ \texttt{Init} \Comment{ initialize networks with pretrained backbone}
\Do
\State $x$, $Y_{\text{gt}}$ $\gets$ \texttt{Sample}($\mathcal{D}$) \Comment{ sample a mini-batch of image and weak-label pairs}
\State $x_s, x_w$ $\gets$ $\mathcal{T}_s(x),\mathcal{T}_w(x)$ \Comment{ apply strong and weak augumentations}
\State $\{x_w^{s}\}$ $\gets$ $ \texttt{multiscale}(x_w)$ \Comment{ generate a set of $x_w$ with different scales}
\State $\{\mathcal{M}^{\prime},~\mathcal{M}^{\dagger\prime},~\mathcal{S}^{\prime}\}$ $\gets$ $\mathcal{F}_{\Theta^{\prime}}(\{x_w^{s}\})$ \Comment{ forward a set of $x_w$ in assignment network}

\State $\mathcal{M}^{\prime},~\mathcal{M}^{\dagger\prime},~\mathcal{S}^{\prime}$ $\gets$ \texttt{Maxpool}$(\{\mathcal{M}^{\prime}\})$,~\texttt{Maxpool}$(\{\mathcal{M}^{\dagger\prime}\})$,~\texttt{Avgpool}$(\{\mathcal{S}^{\prime}\})$  \Comment{ ensemble multiscale outputs}

\State $\mathcal{M}^{\prime},~\mathcal{M}^{\dagger\prime},~\mathcal{S}^{\prime}$ $\gets$ \texttt{Filter}$(\mathcal{M}^{\prime},~\mathcal{M}^{\dagger\prime},~\mathcal{S}^{\prime})$, \Comment{ filter CAMs and segmentation prediction with $Y_{\text{gt}}$}

\State $Z,~Z^{\dagger},~\mathcal{M},~\mathcal{M}^{\dagger},~\mathcal{S}$ $\gets$ $\mathcal{F}_{\Theta}(x_s)$ \Comment{ forward $x_s$ in online network}

\State $\mathcal{L}_{\text{cls}} + \mathcal{L}_{\text{cls}}^{\mathcal{M}^{\dagger}} $ $\gets$ $\mathcal{L}_{\text{cls}}(Z,Y_{\text{gt}}) + \mathcal{L}_{\text{cls}}(Z^{\dagger},Y_{\text{gt}})$ \Comment{ get classification losses for $\mathcal{M}$ and $\mathcal{M}^{\dagger}$ by eq. (1)}
\State $\xi^\star$ $\gets$ solve eq. (6) with $\mathcal{M}^{\prime}$ \Comment{ get dynamic threshold}
\State $\hat{\mathcal{Y}}^{\text{CPL}}$ $\gets$  eq. (2) with $\mathcal{M}^{\prime}, ~ \xi^\star$  \Comment{ obtain CPL}
\State ${\mathcal{P}}$ $\gets$  eq. (4) with $\mathcal{M}^{\prime}, ~ \xi^\star$  \Comment{ estimate perplexity score}
\State ${\mathcal{L}_{\text{c2s}}}$ $\gets$  eq. (5) with $\hat{\mathcal{Y}}^{\text{CPL}}, \mathcal{S}, ~ {\mathcal{P}}$  \Comment{ get CAM2seg loss}
\State ${\mathcal{L}_{\text{c2s}}^{\mathcal{M}^{\dagger}}}$ $\gets$  follow 14 -- 17 but with $\mathcal{M}^{\dagger\prime}$   \Comment{ get another CAM2seg loss}
\State $\hat{\mathcal{Y}}^{\text{SPL}}$ $\gets$  eq. (7) with $\mathcal{S}^{\prime}$  \Comment{ obtain SPL}
\State ${\mathcal{L}_{\text{s2c}}}$ $\gets$  eq. (8) with $\hat{\mathcal{Y}}^{\text{SPL}},~\mathcal{M}$  \Comment{ get Seg2CAM loss}
\State $\mathcal{R}^{+},~\mathcal{R}^{-}$ $\gets$ eq. (9) with $\mathcal{P},~\hat{\mathcal{Y}}^{\text{CPL}}$  \Comment{ define positive and negative correlation matrix}
\State ${\mathcal{L}_{\text{csc}}}$ $\gets$  eq. (10) with $\mathcal{M},~\mathcal{R}^{+},~\mathcal{R}^{-}$  \Comment{ get contrastive seperation loss}
\State $\mathcal{L}_{\text{CoSA}}$ $\gets$ $\! \mathcal{L}_{\text{cls}}\!+ \mathcal{L}_{\text{cls}}^{\mathcal{M}^\dagger}\!\! +\! \lambda_{\text{c2s}}\big(\mathcal{L}_{\text{c2s}}\!+\!\mathcal{L}_{\text{c2s}}^{\mathcal{M}^\dagger}\big)\!+\!\lambda_{\text{s2c}}\mathcal{L}_{\text{s2c}}+\!\lambda_{\text{csc}}\mathcal{L}_{\text{csc}}.$  \Comment{ weighted sum as the  overall training objective}
\State $\Delta\Theta$ $\gets$ $-\nabla_{\mathcal{L}_{\text{CoSA}}}\Theta$ \Comment{ backpropagate the overall loss}
\State $\Theta$ $\gets$ $\Theta+\Delta\Theta$ \Comment{ undate online network with gradient}
\State $\Theta^{\prime}$ $\gets$ $m\Theta^{\prime}+(1-m)\Theta$\Comment{ undate assignment network via EMA}
\doUntil{$\mathcal{L}_{\text{CoSA}}$ converge}
\State \textbf{end}
\end{algorithmic}
\end{algorithm}

\section{Further Implementation Details}
\vspace{5pt}\noindent\textbf{CoSA Implementation Details.}
For image preprocessing, weak transformation $\mathcal{T}_{w}$ and strong transformation $\mathcal{T}_{s}$ are employed in CoSA for the input of assignment network and online network, respectively. $\mathcal{T}_{w}$  and $\mathcal{T}_{s}$ details are given in \cref{tab:weak_aug} and \cref{tab:strong_aug}. Following \cite{ru2023token}, we use the multi-scale inference in assignment network to produce CPL and SPL.
For VOC training, CoSA is warmed up with 6K iterations, where $\lambda_{\text{c2s}}$, $\lambda_{\text{c2s}}$, and $\lambda_{\text{csc}}$ are set to $0$. In practice, we train CoSA for 20K iterations on 2 GPUs, with 2 images per GPU, or for 40K iterations on 1 GPU for some ablation experiments. For COCO training, CoSA is warmed up with 10K iterations and is trained on 2 GPUs, handling 4 images per GPU across 40K iterations.

\vspace{5pt}\noindent\textbf{CoSA-MS Implementation Details.} Tab. 2 in the main paper presents the segmentation results of the multi-stage version of our approach, known as CoSA-MS. In those experiments, we leverage the CAM pseudo-labels generated by our CoSA to \textit{directly} train standalone segmentation networks. It is important to note that we do not use PSA \cite{ahn2018learning}, which is widely used in \cite{xu2022multi,cheng2023out}, nor IRN \cite{ahn2019weakly}, extensively used in \cite{rong2023boundary,chen2023extracting,wang2023treating,kweon2023weakly}, for CPL post-refinement. For our R101 segmentation network, we use a ResNet101 version of DeepLabV3+ model, same as BECO \cite{rong2023boundary}. As for the CoSA-MS with WR38 network, we utilize a encoder-decoder framework, where encoder is WideResNet38 \cite{wu2019wider} and decoder is LargeFoV \cite{chen2017deeplab}, following the final step described in MCT \cite{xu2022multi}.
Regarding the SWIN implementation, we use the SWIN-Base encoder \cite{liu2021swin} in conjunction with UperNet decoder \cite{xiao2018unified}, following the description in \cite{chen2022class,chen2023extracting}.

\vspace{5pt}\noindent\textbf{Training Pseudo Code.} we present the pseudo code for training CoSA in \cref{alg}.

\section{Additional Results}
\subsection{Contribution of each Component}
In addition to the module analysis in Sec. 4.2, we further present additional ablation studies regarding the impact of each component on the baseline. A one-stage WSSS framework, proposed in RRM \cite{zhang2020reliability}, is utilized as the baseline model, albeit without the inclusion of the offline refinement module. As illustrated in \cref{tab:more_module}, it is evident that all components proposed in CoSA exhibit positive effects on the baseline model.

  \begin{table}[ht]
  \addtolength{\tabcolsep}{1.5pt}
      \centering
  \begin{tabular}{cccccclll}
        \toprule

                           &             &             &              &              &             &  & \multicolumn{2}{@{}c@{}}{\textbf{mIoU} @ \texttt{val} \gain{\scriptsize{(increase)}}} \\
    \cmidrule{8-9}
      \textbf{Base.}       & \textbf{GC} & \textbf{SA} & \textbf{RAW} & \textbf{CSC} & \textbf{DT} &  & VOC                                                                   & COCO \\
    \midrule
   \cmark                  &             &             &              &              &             &  & 55.96                                                                 & 37.32  \\
   \cmark                  & \cmark      &             &              &              &             &  & 63.09 \scriptsize{\gain{{(+7.13)}}}                                   & 42.55 \scriptsize{\gain{{(+5.23)}}} \\
   \cmark                  & \cmark      & \cmark      &              &              &             &  & 64.41 \scriptsize{\gain{{(+8.45)}}}                                   & 43.92 \scriptsize{\gain{{(+6.60)}}} \\
         \cmark            &             &             & \cmark       &              &             &  & 67.04 \scriptsize{\gain{{(+11.08)}}}                                  & 42.65 \scriptsize{\gain{{(+5.33)}}} \\
     \cmark                &             &             &              & \cmark       &             &  & 69.24 \scriptsize{\gain{{(+13.27)}}}                                  & 44.11 \scriptsize{\gain{{(+6.79)}}} \\
        \cmark             &             &             &              &              & \cmark      &  & 61.80 \scriptsize{\gain{{(+5.84)}}}                                   & 43.28 \scriptsize{\gain{{(+5.96)}}} \\
        \bottomrule
  \end{tabular}
  \caption{{\textbf{Contribution of Each Component.} We systematically include the proposed components on the baseline (\textbf{Base.}). \textbf{GC}: Guided CAMs, \textbf{SA}: Swapping Assignments, \textbf{RAW}: Reliability based Adaptive Weighting, \textbf{CSC}: Contrastive Separation in CAMs, and \textbf{DT}: Dynamic Threshold.}}
\label{tab:more_module}
\end{table}

\subsection{Hyper-parameter Finetuning}
\noindent we examine the impact of hyper-parameter variation with CoSA resulting from our finetuning. The fine-tuning of each hyper-parameter is demonstrate with the remaining parameters fixed at their determined optimal values. Those hyper-parameters were tuned on VOC \texttt{val} set, and the SOTA results on COCO were achieved with the same hyper-parameters, except batch size and number of iterations were tweaked to accommodate the dataset size difference.

\vspace{5pt}\noindent\textbf{Loss Weights.} We demonstrate the finetuning of the Seg2CAM and CAM2Seg loss weights in \cref{tab:pf_all}(a)(b). A significant mIoU decrease is observed as $\lambda_{c2s}$ reduces the influence of the segmentation branch, as expected. The mIoU reaches its peak when $\lambda_{\text{s2c}}\!=\!0.05$ and $\lambda_{\text{c2s}}\!=\!0.1$.

\vspace{5pt}\noindent\textbf{Low-perplexity Filter.} We finetune the coefficient for the low-pass perplexity filter $\epsilon$, described in eq. (9) of the main paper. The corresponding findings are illustrated in \cref{tab:pf_all}(c). Optimum performance is obtained when $\epsilon$ is set to $1$, either decreasing or increasing this value can impair the performance of our model.

\vspace{5pt}\noindent\textbf{EMA Momentum.} Here, the momentum used for updating the assignment network is finetuned.
Results presented in \cref{tab:pf_all}(d) indicate that the optimal performance is achieved when $m=0.9994$. Additionally, we find that setting $m=1$ freezes the assignment network, breaking the training of online network and leading to framework collapse.

\begin{figure}[ht]
  \centering
        \includegraphics[width=0.5\linewidth]{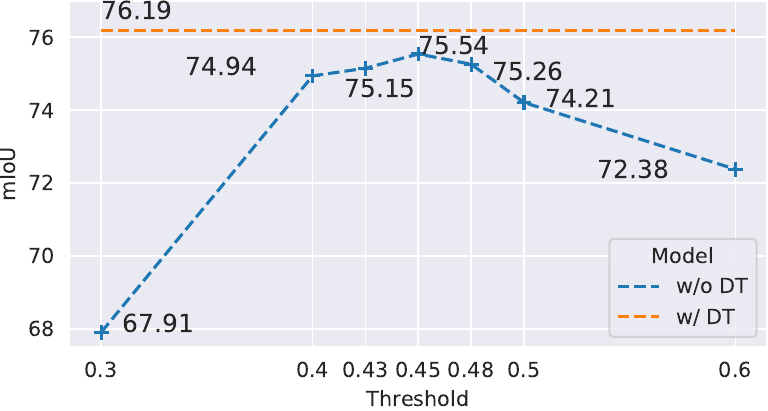}
        \caption{\textbf{Threshold finetuning.}
      (\textit{left}) determined dynamic threshold during training. (\textit{right}) mIoU comparison of fixed threshold \textit{vs.} the purposed dynamic threshold on VOC \texttt{val}.
      }
        \label{fig:AB_DT}
\end{figure}

\vspace{5pt}\noindent\textbf{Fixed Threshold \textit{vs.} Dynamic Threshold.} We evaluate CoSA with predetermined thresholds. The results are presented in \cref{fig:AB_DT}. As shown, the performance peaks when this threshold is set to $0.45$, with an mIoU of $75.54\%$.
However, our dynamic threshold can outperform the best manual finetuning by $0.65\%$. Despite the incurred additional $10\%$ computation overhead, our threshold searching algorithm obviates time-consuming finetuning efforts, resulting in nearly 80\% reduction in hyper-parameter searching time in this case and $(1-1.1n^{-1})\%$ in general where $n$ thresholds are considered. In addition, the adoption of dynamic thresholding can enhance the generalizability to novel datasets.

  \begin{table}[ht]
  \begin{minipage}[t]{0.24\linewidth}
    \begin{center}
      \textbf{{(a)} Seg2CAM weight $\lambda_{\text{s2c}}$} \\
  \small
    \begin{tabular}{c|l}
        \toprule
        $\lambda_{\text{s2c}}$ & \textbf{mIoU} \\
         \midrule
        $0.2$ & 73.79 \\
        $0.1$ & 74.67 \\
        $0.05$ & \textbf{76.19} \\
        $0.025$ & 75.25 \\
        $0.0125$ & 74.33 \\
        \bottomrule
    \end{tabular}
    \end{center}
\end{minipage}
\hfill
  \begin{minipage}[t]{0.24\linewidth}
    \begin{center}
      \textbf{{(b)} CAM2Seg weight $\lambda_{\text{c2s}}$} \\
  \small
    \begin{tabular}{c|l}
        \toprule
        $\lambda_{\text{c2s}}$ & \textbf{mIoU} \\
         \midrule
        $0.4$ & 74.67 \\
        $0.2$ & 75.56 \\
        $0.1$ & \textbf{76.19} \\
        $0.05$ & 73.95 \\
        $0.025$ & 61.55 \\
        \bottomrule
                  \end{tabular}
    \end{center}
\end{minipage}
  \begin{minipage}[t]{0.24\linewidth}
    \begin{center}
      \textbf{(c) Perplexity filter $\epsilon$} \\
  \small
    \begin{tabular}{c|l}
        \toprule
         $\epsilon$ & \textbf{mIoU} \\
         \midrule
        $\infty$ & 73.66 \\
        $2$ & 75.52 \\
        $1$ & \textbf{76.19} \\
        $0.5$ & 74.30 \\
        $0.1$ & 70.63 \\
        \bottomrule
                  \end{tabular}
    \end{center}
\end{minipage}
\hfill
  \begin{minipage}[t]{0.24\linewidth}
    \begin{center}
      \textbf{{(d)} Momentum $m$} \\
  \small
    \begin{tabular}{c|l}
        \toprule
         $m$ & \textbf{mIoU} \\
         \midrule
         $0.9990$ & 73.79 \\
         $0.9992$ & 75.40 \\
        $0.9994$ & \textbf{76.19}  \\
        $0.9996$ & 75.58  \\
        $0.9999$ &  71.42 \\
        $1.0000$ &  15.99 \\
        \bottomrule
                  \end{tabular}
    \end{center}
\end{minipage}
\caption{\textbf{Hyper-parameter Finetuning Results.}
  Parameter searching for  \textbf{(a)} Loss weight for CAM2Seg $\lambda_{\text{c2s}}$; \textbf{(b)} Loss weight for Seg2CAM $\lambda_{\text{s2c}}$; \textbf{(c)} Low-pass perplexity filter coefficient $\epsilon$;
  \textbf{(e)} EMA Momentum $m$ for updating assignment network.
  \textbf{mIoU} represents semantic segmentation result on PASCAL VOC \texttt{val} split.
}
\label{tab:pf_all}
      \end{table}


\subsection{Per-class Segmentation Comparisons}
We show the per-class semantic segmentation results on VOC \texttt{val} and \texttt{test} splits as well as COCO \texttt{val} split.

\vspace{5pt}\noindent\textbf{Comparisons on VOC.}
\cref{tab:percls_voc_val} illustrates the CoSA per-class mIoU results compared with recent works: AdvCAM \cite{lee2022anti}, MCT \cite{xu2022multi}, ToCo \cite{ru2023token}, Xu \etal \cite{xu2023learning}, BECO \cite{rong2023boundary}. To be fair in comparison, we include CoSA with CRF \cite{chen2017deeplab} postprocessing results, denoted as CoSA$^*$, same as other SOTA models.
Notably, CoSA dominates in 10 out of 21 classes. In particular, categories like boat ($5.9\%\!\uparrow$), chair ($8.2\%\!\uparrow$), and sofa ($17.2\%\!\uparrow$), demonstrate substantial lead over the SOTA models. In the VOC \texttt{test} split (depicted in \cref{tab:percls_voc_test}), we still observe its superiority over other SOTA methods, where CoSA dominates in 15 out of 21 classes.

\vspace{5pt}\noindent\textbf{Comparisons on COCO.}
We compare CoSA with recent WSSS works for individual class performance on the COCO \texttt{val} set. As illustrated in \cref{tab:percls_coco}, CoSA outperforms its counterparts in 56 out of 81 classes. Particularly, classes such as truck ($10.6\%\!\uparrow$), tie ($14.3\%\!\uparrow$), kite ($12.4\%\!\uparrow$), baseball glove ($20.3\%\!\uparrow$), knife ($14.5\%\!\uparrow$), ($10.6\%\!\uparrow$), carrot ($13.0\%\!\uparrow$), donuts ($10.0\%\!\uparrow$), couch ($13.9\%\!\uparrow$), oven ($13.0\%\!\uparrow$), and toothbrush ($10.0\%\!\uparrow$) exhibit remarkable leading performance.

\begin{table}[ht]
    \renewcommand\arraystretch{1}
    \begin{minipage}[t]{1\textwidth}
    \small
    \centering
\resizebox{1\columnwidth}{!}{
    \begin{tabular}{l|ccccccccccc}
    \toprule
	Method                                       & bkg           & plane         & bike          & bird          & boat          & bottle        & bus           & car           & cat           & chair         & cow \\
	\midrule
	AdvCAM \cite{lee2021anti} \tiny{CVPR21}      & 90.0          & 79.8          & 34.1          & 82.6          & 63.3          & 70.5          & 89.4          & 76.0          & 87.3          & 31.4          & 81.3\\
	MCT \cite{xu2022multi} \tiny{CVPR22}         & 91.9          & 78.3          & 39.5          & \textbf{89.9} & 55.9          & 76.7          & 81.8          & 79.0          & 90.7          & 32.6          & 87.1 \\
	ToCo \cite{ru2023token} \tiny{CVPR23}        & 91.1          & 80.6          & \textbf{48.7} & 68.6          & 45.4          & \textbf{79.6} & 87.4          & 83.3          & 89.9          & 35.8          & 84.7 \\
	Xu \etal \cite{xu2023learning} \tiny{CVPR23} & 92.4          & 84.7          & 42.2          & 85.5          & 64.1          & 77.4          & 86.6          & 82.2          & 88.7          & 32.7          & 83.8 \\
	BECO \cite{rong2023boundary} \tiny{CVPR23}   & 91.1          & 81.8          & 33.6          & 87.0          & 63.2          & 76.1          & \textbf{92.3} & \textbf{87.9} & \textbf{90.9} & 39.0          & \textbf{90.2} \\
	CoSA* (Ours)                                  & \textbf{93.1} & \textbf{85.5} & 48.5          & 88.7          & \textbf{70.0} & 77.6          & 90.4          & 86.4          & 90.3          & \textbf{47.2} & 88.7\\
	\midrule
	Method                                       & table         & dog           & horse         & mbike         & person        & plant         & sheep         & sofa          & train         & tv            & \textbf{mIoU}\\
	\midrule
	AdvCAM \cite{lee2021anti} \tiny{CVPR21}      & 33.1          & 82.5          & 80.8          & 74.0          & 72.9          & 50.3          & 82.3          & 42.2          & \textbf{74.1} & 52.9          & 68.1 \\
	MCT \cite{xu2022multi} \tiny{CVPR22}         & 57.2          & 87.0          & 84.6          & 77.4          & 79.2          & 55.1          & 89.2          & 47.2          & 70.4          & 58.8          & 71.9 \\
	ToCo \cite{ru2023token} \tiny{CVPR23}        & \textbf{60.5} & 83.7          & 83.7          & 76.8          & 83.0          & 56.6          & 87.9          & 43.5          & 60.5          & 63.1          & 71.1 \\
	Xu \etal \cite{xu2023learning} \tiny{CVPR23} & 59.0          & 82.4          & 80.9          & 76.1          & 81.4          & 48.0          & 88.2          & 46.4          & 70.2          & 62.5          & 72.2 \\
	BECO \cite{rong2023boundary} \tiny{CVPR23}   & 41.6          & 85.9          & 86.3          & \textbf{81.8} & 76.7          & \textbf{56.7} & 89.5          & 54.7          & 64.3          & 60.6          & 72.9 \\
	CoSA* (Ours)                                  & 54.1          & \textbf{87.3} & \textbf{87.1} & 79.6          & \textbf{85.6} & 53.2          & \textbf{89.9} & \textbf{71.9} & 65.1          & \textbf{63.4} & \textbf{76.4} \\
	\bottomrule
    \end{tabular}
}
    \caption{\textbf{Per-class Segmentation on VOC \texttt{val} Split.}
    Comparison of per-class segmentation results on VOC \texttt{val}. CoSA is compared with AdvCAM, MCTformer, ToCo, Xu \textit{et al.} and BECO. Best results are in \textbf{bold}.
}
    \label{tab:percls_voc_val}
\end{minipage}
\\[30pt]
\begin{minipage}[t]{1\textwidth}
    \small
    \centering
\resizebox{1\columnwidth}{!}{
    \begin{tabular}{l|ccccccccccc}
    \toprule
	Method                                    & bkg           & plane         & bike          & bird          & boat          & bottle        & bus           & car           & cat           & chair         & cow \\
	\midrule
	AdvCAM   \cite{lee2021anti} \tiny{CVPR21} & 90.1          & 81.2          & 33.6          & 80.4          & 52.4          & 66.6          & 87.1          & 80.5          & 87.2          & 28.9          & 80.1 \\
	MCT \cite{xu2022multi} \tiny{CVPR22}      & 90.9          & 76.0          & 37.2          & 79.1          & 54.1          & 69.0          & 78.1          & 78.0          & 86.1          & 30.3          & 79.5\\
	ToCo \cite{ru2023token} \tiny{CVPR23}     & 91.5          & \textbf{88.4} & \textbf{49.5} & 69.0          & 41.6          & 72.5          & 87.0          & 80.7          & 88.6          & 32.2          & 85.0 \\
	CoSA* (Ours)                               & \textbf{93.3} & 88.1          & 47.0          & \textbf{84.2} & \textbf{60.2} & \textbf{75.0} & \textbf{87.7} & \textbf{81.7} & \textbf{92.0} & \textbf{34.5} & \textbf{87.8} \\
	\midrule
	Method                                    & table         & dog           & horse         & mbike         & person        & plant         & sheep         & sofa          & train         & tv            & \textbf{mIoU}\\
	\midrule
	AdvCAM   \cite{lee2021anti} \tiny{CVPR21} & 38.5          & 84.0          & 83.0          & 79.5          & 71.9          & 47.5          & 80.8          & 59.1          & 65.4          & 49.7          & 68.0 \\
	MCT \cite{xu2022multi} \tiny{CVPR22}      & 58.3          & 81.7          & 81.1          & 77.0          & 76.4          & 49.2          & 80.0          & 55.1          & 65.4          & 54.5          & 68.4 \\
	ToCo \cite{ru2023token} \tiny{CVPR23}     & \textbf{68.4} & 81.4          & 85.6          & 83.2          & \textbf{83.4} & \textbf{68.2} & \textbf{88.9} & 55.0          & 49.3          & \textbf{65.0} & 72.2 \\
	CoSA* (Ours)                               & 59.6          & \textbf{86.2} & \textbf{86.3} & \textbf{84.9} & 82.8          & \textbf{68.2} & 87.4          & \textbf{63.9} & \textbf{67.7} & 61.6          & \textbf{75.2} \\
	\bottomrule
    \end{tabular}
}
    \caption{\textbf{Per-class Segmnetation on VOC \texttt{test} Split.}
    Comparison of per-class segmentation results on VOC \texttt{test}. Results from AdvCAM, MCT, and ToCo are used for this comparison. Best results are in \textbf{bold}. }
    \label{tab:percls_voc_test}

\end{minipage}
\end{table}

\begin{table}[ht]
    \renewcommand\arraystretch{1.1}
    \normalsize
    \centering
\resizebox{1\columnwidth}{!}{
    \begin{tabular}{m{52pt} m{35pt}<{\centering} m{45pt}<{\centering} m{35pt}<{\centering} m{30pt}<{\centering} | m{50pt} m{35pt}<{\centering} m{45pt}<{\centering} m{35pt}<{\centering} m{30pt}<{\centering}}
    \toprule
    Class          & MCT\scriptsize\cite{xu2022multi} \tiny{(CVPR22)}          & Xu \etal\scriptsize\cite{xu2023learning} \tiny{(CVPR23)}    & ToCo\scriptsize\cite{ru2023token} \tiny{(CVPR23)} & CoSA* \tiny{(Ours)}   & Class          & MCT\scriptsize\cite{xu2022multi} \tiny{(CVPR22)}          & Xu \etal\scriptsize\cite{xu2023learning} \tiny{(CVPR23)}    & ToCo\scriptsize\cite{ru2023token} \tiny{(CVPR23)} & CoSA* \tiny{(Ours)}  \\
    \midrule
    background     & 82.4          & \textbf{85.3} & 68.5 & 84.0          & wine glass    & 27.0 & 33.8          & 20.6          & \textbf{42.1} \\
    person         & 62.6          & \textbf{72.9} & 28.1 & 70.3          & cup           & 29.0 & \textbf{35.8} & 26.0          & 33.1 \\
    bicycle        & 47.4          & 49.8          & 39.7 & \textbf{52.4} & fork          & 23.4 & 20.0          & 7.6           & \textbf{24.2} \\
    car            & 47.2          & 43.8          & 38.9 & \textbf{54.3} & knife         & 12.0 & 12.6          & 18.4          & \textbf{32.9} \\
    motorcycle     & 63.7          & 66.2          & 55.1 & \textbf{71.9} & spoon         & 6.6  & 6.7           & 3.0           & \textbf{9.0}  \\
    airplane       & 64.7          & 69.2          & 62.1 & \textbf{74.0} & bowl          & 22.4 & \textbf{23.7} & 19.8          & 22.8 \\
    bus            & 64.5          & 69.1          & 39.0 & \textbf{77.2} & banana        & 63.2 & 64.4          & \textbf{71.5} & 69.3 \\
    train          & \textbf{64.5} & 63.7          & 48.7 & 60.0          & apple         & 44.4 & 50.8          & 55.5          & \textbf{61.3} \\
    truck          & 44.8          & 43.4          & 37.3 & \textbf{55.4} & sandwich      & 39.7 & 47.0          & 41.2          & \textbf{48.3} \\
    boat           & 42.3          & 42.3          & 49.1 & \textbf{52.1} & orange        & 63.0 & 64.6          & \textbf{70.6} & 69.2 \\
    traffic light  & 49.9          & 49.3          & 47.3 & \textbf{55.1} & broccoli      & 51.2 & 50.6          & \textbf{56.7} & 52.8 \\
    fire hydrant   & 73.2          & 74.9          & 69.6 & \textbf{78.8} & carrot        & 40.0 & 38.6          & 46.4          & \textbf{59.4} \\
    stop sign      & 76.6          & 77.3          & 70.1 & \textbf{82.2} & hot dog       & 53.0 & 54.0          & \textbf{60.1}          & 59.9 \\
    park meter  & 64.4          & 67.0          & 67.9 & \textbf{71.5} & pizza         & 62.2 & \textbf{64.1} & 54.9          & 56.5 \\
    bench          & 32.8          & 34.1          & 43.9 & \textbf{50.2} & donut         & 55.7 & 59.7          & 61.1          & \textbf{71.1} \\
    bird           & 62.6          & 63.1          & 58.6 & \textbf{65.4} & cake          & 47.9 & 50.6          & 42.5          & \textbf{57.0} \\
    cat            & 78.2          & 76.2          & 74.0 & \textbf{79.8} & chair         & 22.8 & 24.5          & 24.1          & \textbf{33.8} \\
    dog            & 68.2          & 70.6          & 64.0 & \textbf{72.8} & couch         & 35.0 & 40.0          & 44.2          & \textbf{58.1} \\
    horse          & 65.8          & 67.1          & 66.1 & \textbf{71.4} & plant  & 13.5 & 13.0          & \textbf{27.4} & 23.5 \\
    sheep          & 70.1          & 70.8          & 67.9 & \textbf{74.3} & bed           & 48.6 & 53.7          & 54.0          & \textbf{61.5} \\
    cow            & 68.3          & 71.2          & 69.0 & \textbf{74.0} & table  & 12.9 & 19.2          & 25.6          & \textbf{29.2} \\
    elephant       & 81.6          & \textbf{82.2} & 79.7 & 81.9          & toilet        & 63.1 & 66.6          & 62.0          & \textbf{69.7} \\
    bear           & 80.1          & 79.6          & 76.8 & \textbf{85.3} & tv            & 47.9 & 50.8          & 49.1          & \textbf{53.2} \\
    zebra          & \textbf{83.0} & 82.8          & 77.5 & 76.3          & laptop        & 49.5 & 55.4          & 55.7          & \textbf{63.9} \\
    giraffe        & \textbf{76.9} & 76.7          & 66.1 & 68.5          & mouse         & 13.4 & 14.4          & 8.6           & \textbf{16.4} \\
    backpack       & 14.6          & 17.5          & 20.3 & \textbf{28.6} & remote        & 41.9 & 47.1          & \textbf{56.6} & 49.1 \\
    umbrella       & 61.7          & 66.9          & 70.9 & \textbf{73.4} & keyboard      & 49.8 & \textbf{57.2} & 41.8          & 49.6 \\
    handbag        & 4.5           & 5.8           & 8.1  & \textbf{11.9} & cellphone     & 54.1 & 54.9          & 58.5          & \textbf{66.2} \\
    tie            & 25.2          & 31.4          & 33.4 & \textbf{47.7} & microwave     & 38.0 & 46.1          & \textbf{55.5} & 53.2 \\
    suitcase       & 46.8          & 51.4          & 55.3 & \textbf{63.8} & oven          & 29.9 & 35.3          & 36.2          & \textbf{49.2} \\
    frisbee        & 43.8          & 54.1          & 39.6 & \textbf{63.1} & toaster       & 0.0  & \textbf{2.0}  & 0.0           & 0.0  \\
    skis           & 12.8          & 13.0          & 4.0  & \textbf{22.5} & sink          & 28.0 & 36.1          & 19.0          & \textbf{41.9} \\
    snowboard      & 31.4          & 30.3          & 15.5 & \textbf{40.5} & refrigerator  & 40.1 & 52.7          & 51.9          & \textbf{62.0} \\
    sports ball    & 9.2           & \textbf{36.1} & 11.0 & 33.1          & book          & 32.2 & 34.8          & 31.5          & \textbf{37.8} \\
    kite           & 26.3          & 47.5          & 40.7 & \textbf{59.9} & clock         & 43.2 & 51.5          & 32.9          & \textbf{55.2} \\
    baseball bat   & 0.9           & \textbf{7.0}  & 1.8  & 3.8           & vase          & 22.6 & 25.8          & 33.3          & \textbf{33.8} \\
    glove & 0.7           & 10.4          & 17.6 & \textbf{37.9} & scissors      & 32.9 & 30.7          & 49.8          & \textbf{54.7} \\
    skateboard     & 7.8           & \textbf{15.2} & 13.3 & 12.5          & teddy bear    & 61.9 & 61.4          & 67.5          & \textbf{69.3} \\
    surfboard      & 46.5          & \textbf{51.5} & 21.5 & 16.5          & hair drier    & 0.0  & 1.3           & \textbf{10.0} & 0.3  \\
    racket  & 1.4           & \textbf{26.4} & 6.8  & 7.2           & toothbrush    & 12.2 & 19.0          & 29.3          & \textbf{39.3} \\ \cline{6-10}
    bottle         & 31.1          & \textbf{37.1} & 25.7 & 35.1          & \textbf{mIoU} & 42.0 & 45.9          & 42.4          & \textbf{51.1} \\
	\bottomrule
    \end{tabular}
}
    \caption{\textbf{Per-class Segmentation Results on COCO.}
    Comparison of per-class segmentation results on COCO \texttt{val}. CoSA is compared with MCT, Xu \textit{et al.} and ToCo. Best results are in \textbf{bold}. }
    \label{tab:percls_coco}
\end{table}

\clearpage

\subsection{Further Qualitative Comparisons}
\begin{figure}[ht]
  \centering
  \includegraphics[width=1\textwidth]{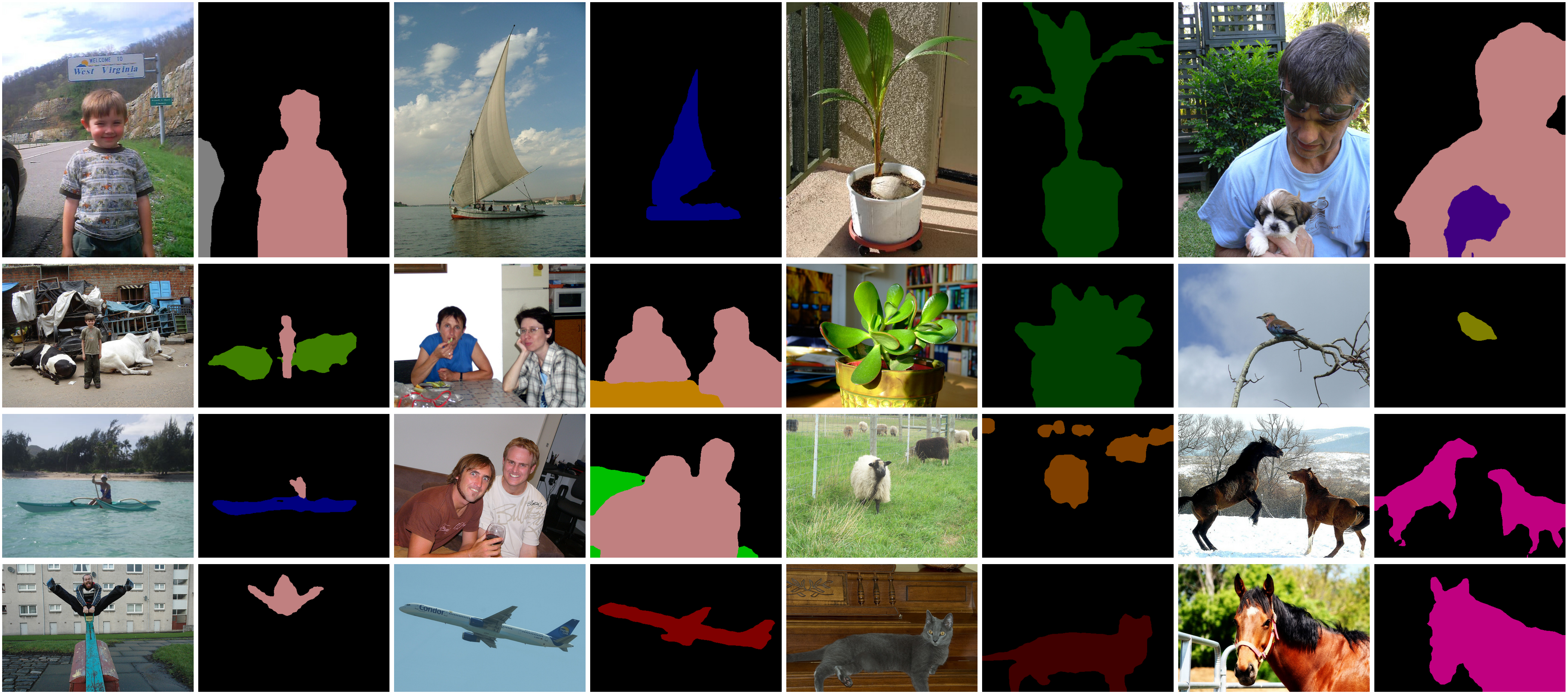}
  \caption{\textbf{Visualization on VOC \texttt{test}.}
    Different colors represent different categories:
    black: background;
    {\protect\tikz \fill[fill={rgb,255:red,128; green,128; blue,128}] (1ex,1ex) circle (1ex);}: car;
    {\protect\tikz \fill[fill={rgb,255:red,192; green,128; blue,128}] (1ex,1ex) circle (1ex);}: person;
    {\protect\tikz \fill[fill={rgb,255:red,0; green,0; blue,128}] (1ex,1ex) circle (1ex);}: boat;
    {\protect\tikz \fill[fill={rgb,255:red,0; green,64; blue,0}] (1ex,1ex) circle (1ex);}: plant;
    {\protect\tikz \fill[fill={rgb,255:red,64; green,0; blue,128}] (1ex,1ex) circle (1ex);}: dog;
    {\protect\tikz \fill[fill={rgb,255:red,64; green,128; blue,0}] (1ex,1ex) circle (1ex);}: cow.
    {\protect\tikz \fill[fill={rgb,255:red,192; green,128; blue,0}] (1ex,1ex) circle (1ex);}: dining-table.
    {\protect\tikz \fill[fill={rgb,255:red,128; green,128; blue,0}] (1ex,1ex) circle (1ex);}: bird;
    {\protect\tikz \fill[fill={rgb,255:red,0; green,192; blue,0}] (1ex,1ex) circle (1ex);}: sofa;
    {\protect\tikz \fill[fill={rgb,255:red,128; green,64; blue,0}] (1ex,1ex) circle (1ex);}: sheep;
    {\protect\tikz \fill[fill={rgb,255:red,192; green,0; blue,128}] (1ex,1ex) circle (1ex);}: house;
    {\protect\tikz \fill[fill={rgb,255:red,128; green,0; blue,0}] (1ex,1ex) circle (1ex);}: airplane;
    {\protect\tikz \fill[fill={rgb,255:red,64; green,0; blue,0}] (1ex,1ex) circle (1ex);}: cat.
  }
  \label{fig:voc_viz_test}
\end{figure}

\begin{figure*}[ht]
  \centering
  \includegraphics[width=1\textwidth]{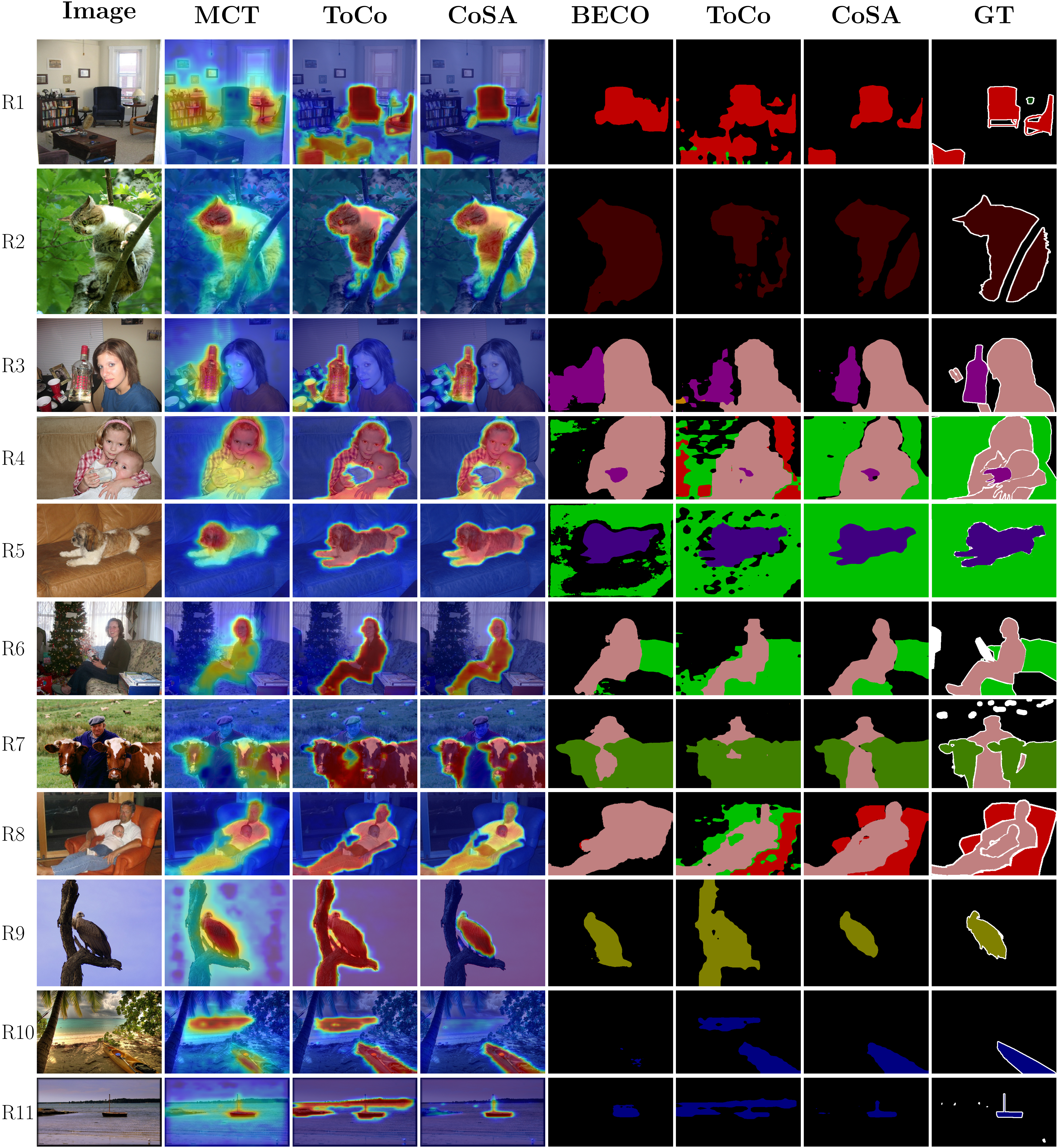}
  \caption{\textbf{Qualitative Comparisons on VOC Dataset.}
    CoSA exhibits
    1) better foreground-background separation (\textit{R1}--\textit{R3});
    2) more robust to inter-class variation and occlusion (\textit{R4}--\textit{R7});
    3) limitations in the ground troth annotations (\textit{R8});
    4) less coexistence problem (\textit{R9}--\textit{R11}).
    Different colors represent different categories:
    black: background; white: ignore areas;
    {\protect\tikz \fill[fill={rgb,255:red,192; green,0; blue,0}] (1ex,1ex) circle (1ex);}: chair;
    {\protect\tikz \fill[fill={rgb,255:red,0; green,64; blue,0}] (1ex,1ex) circle (1ex);}: plant;
    {\protect\tikz \fill[fill={rgb,255:red,64; green,0; blue,0}] (1ex,1ex) circle (1ex);}: cat;
    {\protect\tikz \fill[fill={rgb,255:red,192; green,128; blue,128}] (1ex,1ex) circle (1ex);}: person;
    {\protect\tikz \fill[fill={rgb,255:red,128; green,0; blue,128}] (1ex,1ex) circle (1ex);}: bottle;
    {\protect\tikz \fill[fill={rgb,255:red,0; green,192; blue,0}] (1ex,1ex) circle (1ex);}: sofa;
    {\protect\tikz \fill[fill={rgb,255:red,64; green,0; blue,128}] (1ex,1ex) circle (1ex);}: dog;
    {\protect\tikz \fill[fill={rgb,255:red,64; green,128; blue,0}] (1ex,1ex) circle (1ex);}: cow.
    {\protect\tikz \fill[fill={rgb,255:red,128; green,128; blue,0}] (1ex,1ex) circle (1ex);}: bird;
    {\protect\tikz \fill[fill={rgb,255:red,0; green,0; blue,128}] (1ex,1ex) circle (1ex);}: boat;
    The activated classes in the demonstration from top to bottom are: chair,  cat, bottle, person, dog, person, cow, person, bird, boat, boat.
  }
  \label{fig:voc_cmp_addition}
\end{figure*}

\begin{figure*}[ht]
  \centering
  \includegraphics[width=1\textwidth]{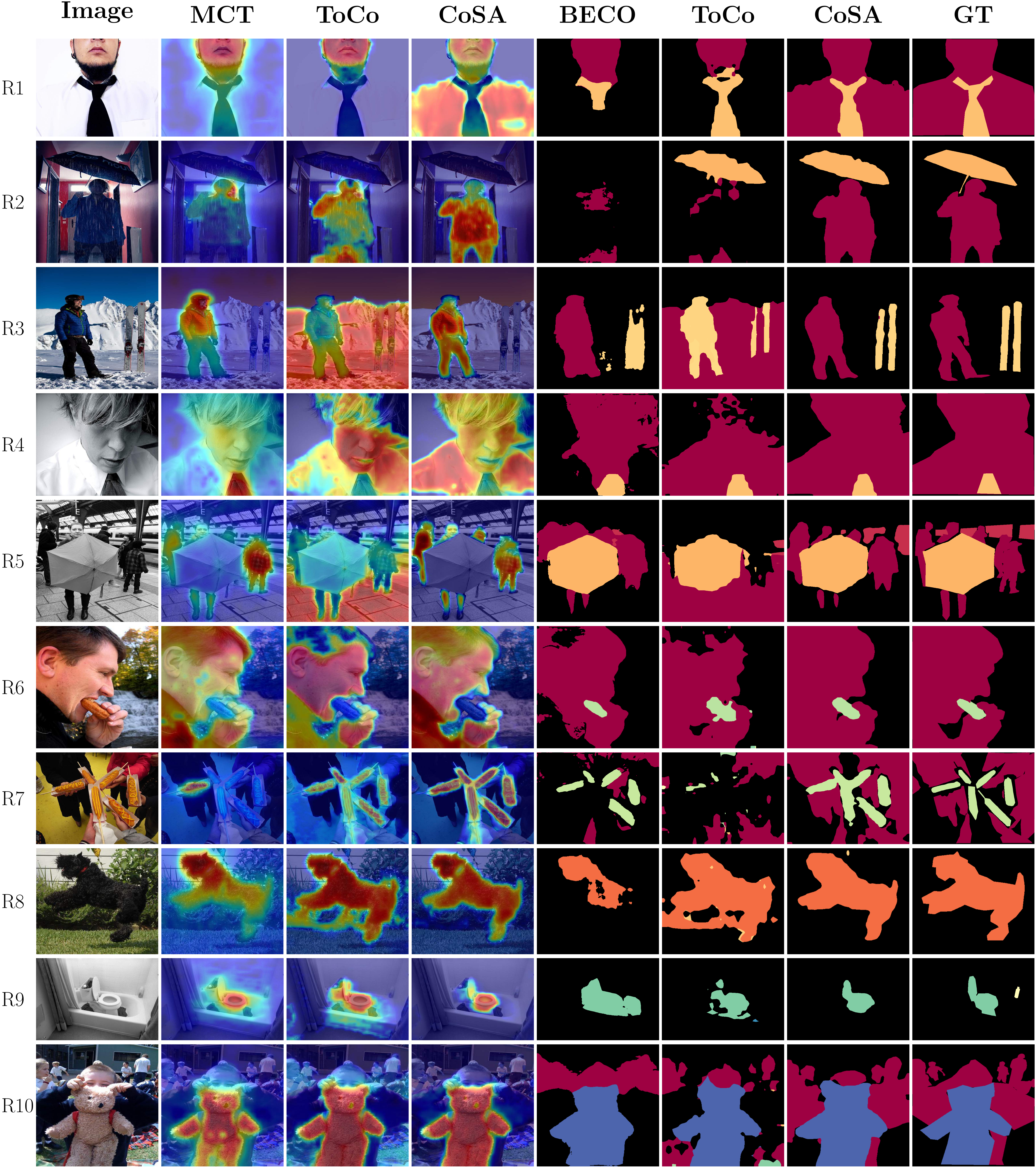}
  \caption{\textbf{Qualitative Comparisons on COCO Dataset.}
    CoSA demonstrates superior quality in terms of foreground-background separation (\textit{R1}--\textit{R10}).
    Categories involved --
    \textit{R1}: \textbf{person}, tie;
  \textit{R2}: \textbf{person}, umbrella;
\textit{R3}: \textbf{person}, skis;
\textit{R4}: \textbf{person}, tie;
\textit{R5}: \textbf{person}, train, umbrella;
\textit{R6}: \textbf{person}, hot dog;
\textit{R7}: person, \textbf{hot dog};
\textit{R8}: \textbf{dog}, frisbee;
\textit{R9}: bottle, \textbf{toilet};
\textit{R10}: person, \textbf{teddy bear};
Categories in \textbf{Bold} denotes the activated classes in CAMs.
}
  \label{fig:coco_cmp_addition1}
\end{figure*}

\begin{figure*}[ht]
  \centering
  \includegraphics[width=1\textwidth]{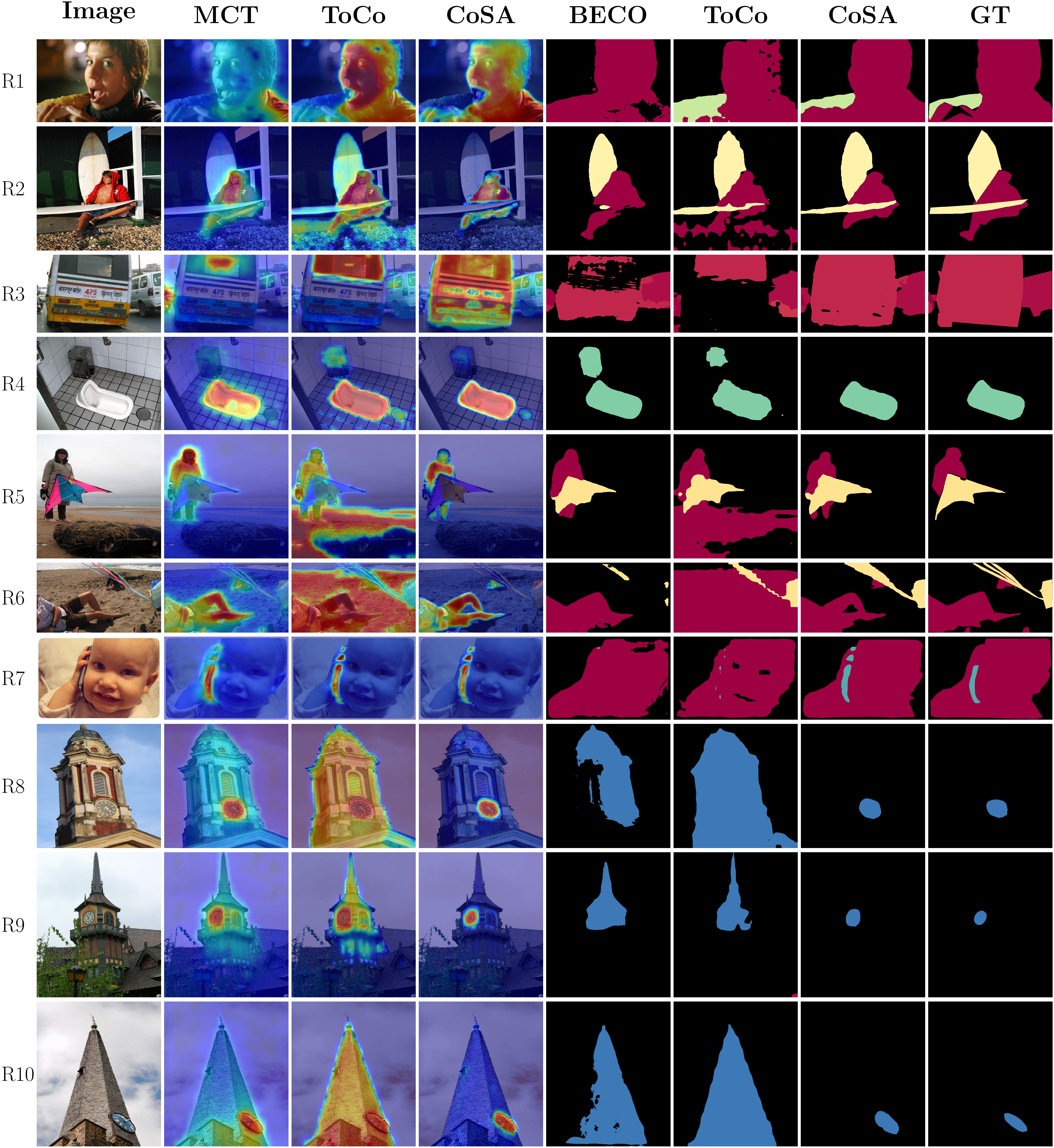}
  \caption{\textbf{More Qualitative Comparisons on COCO Dataset.}
    CoSA shows
    1) more robust to inter-class variation and occlusion (\textit{R1}--\textit{R4});
    2) limitations in the ground troth annotations (\textit{R5}--\textit{R7});
    3) less coexistence problem (\textit{R8}--\textit{R10}).
    Categories involved --
\textit{R1}: \textbf{person}, donuts;
\textit{R2}: \textbf{person}, surfboard.
\textit{R3}: person, car, motorcycle, \textbf{bus};
\textit{R4}: \textbf{toilet};
\textit{R5}: \textbf{person}, kite;
\textit{R6}: \textbf{person}, kite;
\textit{R7}: person, \textbf{cell phone};
\textit{R8}: \textbf{clock};
\textit{R9}: \textbf{clock};
\textit{R10}: \textbf{clock}.
Categories in \textbf{Bold} denotes the activated classes in CAMs.
  }
  \label{fig:coco_cmp_addition2}
\end{figure*}

\begin{figure}[ht]
  \centering
  \includegraphics[width=1\textwidth]{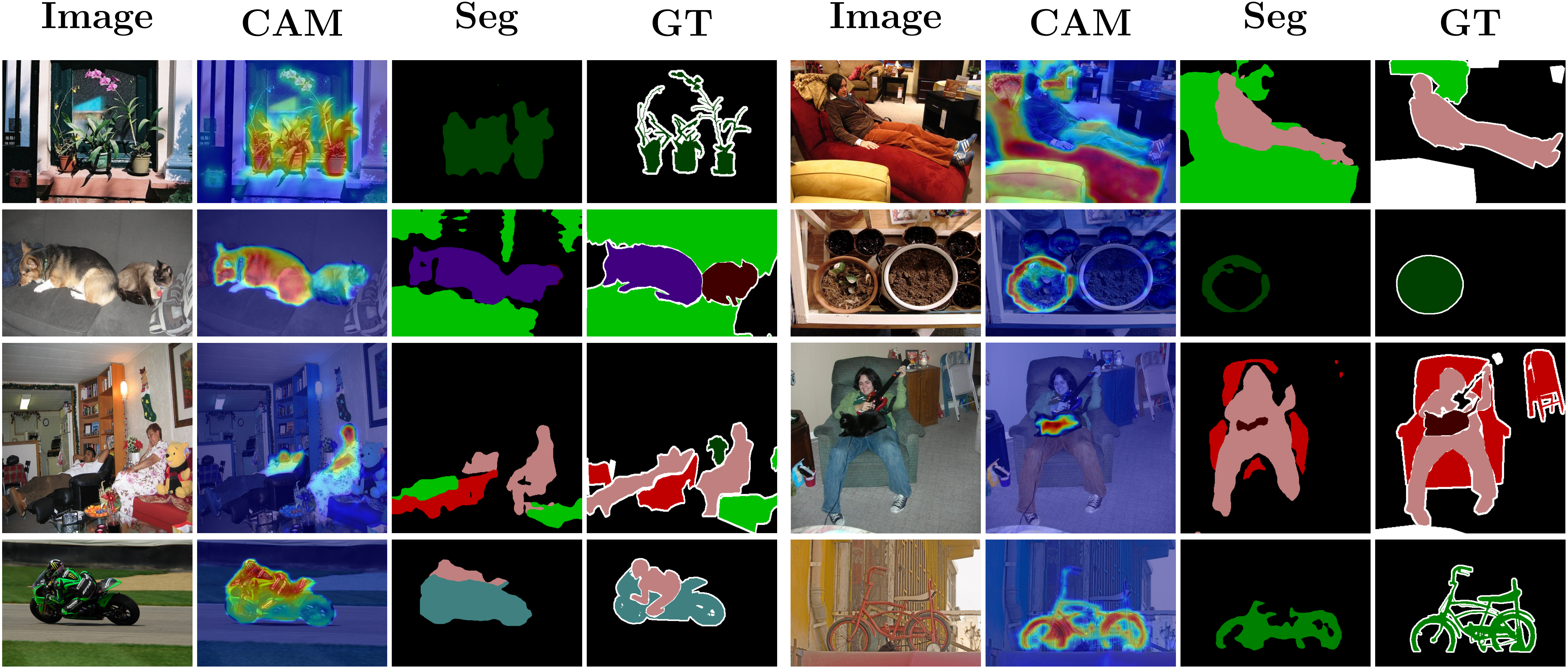}
  \caption{\textbf{Illustrations of CoSA failure Cases.}
    Different colors represent different categories:
    black: background; white: ignore areas;
    {\protect\tikz \fill[fill={rgb,255:red,0; green,64; blue,0}] (1ex,1ex) circle (1ex);}: plant;
    {\protect\tikz \fill[fill={rgb,255:red,192; green,128; blue,128}] (1ex,1ex) circle (1ex);}: person;
    {\protect\tikz \fill[fill={rgb,255:red,0; green,192; blue,0}] (1ex,1ex) circle (1ex);}: sofa;
    {\protect\tikz \fill[fill={rgb,255:red,64; green,0; blue,128}] (1ex,1ex) circle (1ex);}: dog;
    {\protect\tikz \fill[fill={rgb,255:red,64; green,0; blue,0}] (1ex,1ex) circle (1ex);}: cat;
    {\protect\tikz \fill[fill={rgb,255:red,192; green,0; blue,0}] (1ex,1ex) circle (1ex);}: chair;
    {\protect\tikz \fill[fill={rgb,255:red,64; green,128; blue,128}] (1ex,1ex) circle (1ex);}: motorbike;
    {\protect\tikz \fill[fill={rgb,255:red,0; green,128; blue,0}] (1ex,1ex) circle (1ex);}: bicycle.
    The activated classes in the demonstration from left to right and from top to bottom are: plant, sofa, dog, plant, person, cat, person, bicycle.
  }
  \label{fig:voc_viz_fail}
\end{figure}

More visualizations of our CoSA results are given in \cref{fig:voc_cmp_addition} for VOC and \cref{fig:coco_cmp_addition1}, \cref{fig:coco_cmp_addition2} for COCO.
When compared to other SOTA models, CoSA exhibits
i) better foreground-background separation (evidenced in \textit{R2}--\textit{R3} in \cref{fig:voc_cmp_addition} and \textit{R1}--\textit{R10} in \cref{fig:coco_cmp_addition1});
ii) more robust to inter-class variation and occlusion (affirmed in \textit{R4}--\textit{R7} in \cref{fig:voc_cmp_addition} and \textit{R1}--\textit{R4} in \cref{fig:coco_cmp_addition2}).
iii) less coexistence problem (demonstrated in \textit{R9}--\textit{R11} in \cref{fig:voc_cmp_addition} and \textit{R8}-- \textit{R10} in \cref{fig:coco_cmp_addition2});
Last but not least, our CoSA can reveal certain limitations in manual GT segmentation, as depicted in \textit{R8} in \cref{fig:voc_cmp_addition} and \textit{R5}--\textit{R7} in \cref{fig:coco_cmp_addition2}.
We also show our CoSA results on VOC \texttt{test} set in \cref{fig:voc_viz_test} and some failure cases in \cref{fig:voc_viz_fail}.

\clearpage
\clearpage
\clearpage

\clearpage

\begin{table}[ht]
    \begin{minipage}[t]{1\linewidth}
    \centering

\resizebox{1\columnwidth}{!}{
    \begin{tabular}{c|l|l}
    \toprule
        Transformation & Description  &  Parameter Setting\\ \midrule
         RandomRescale & Rescale the image by $r$ times, $r$ randomly sampled from  $r \sim U(r_{min}, r_{max})$. & {$r_{min} = 0.5, r_{max} = 2$ } \\
         RandomFlip&Randomly horizontally flip a image with probability of $p$.& $p=0.5$ \\
         RandomCrop&Randomly crop a image by a hight $h$ and a width $w$. & $w=448, h=448$ \\
         GaussianBlur&Randomly blur a image with probability of $p$. & $p=0.5$ \\
         \bottomrule
    \end{tabular}
}
    \caption{Weak data augmentation $\mathcal{T}_{w}$ for the input of assignment network.}
    \label{tab:weak_aug}
\end{minipage} \\

    \begin{minipage}[t]{1\linewidth}
    \centering
\resizebox{1\columnwidth}{!}{
    \begin{tabular}{c|l|l}
    \toprule
        Transformation & Description  &  Parameter Setting\\ \midrule
         RandomRescale & Rescale the image by $r$ times, $r$ randomly sampled from  $r \sim U(r_{min}, r_{max})$. & {$r_{min} = 0.5, r_{max} = 2$ } \\
         RandomFlip&Randomly horizontally flip a image with probability of $p$.& $p=0.5$ \\
         RandomCrop&Randomly crop a image by a hight $h$ and a width $w$. & $w=448, h=448$ \\
         GaussianBlur&Randomly blur a image with probability of $p$. & $p=0.5$ \\
         OneOf & Select one of the transformation in a transformation set $T$.& $T=$  \texttt{TransAppearance}\\
         \bottomrule
    \end{tabular}
}
    \caption{Strong data augmentation $\mathcal{T}_{s}$ for the input of online network image.}
    \label{tab:strong_aug}
\end{minipage}
\\
    \begin{minipage}[t]{1\linewidth}
    \centering
\resizebox{1\columnwidth}{!}{
    \begin{tabular}{c|l|l}
    \toprule
        Transformation & Description  &  Parameter Setting\\
        \midrule
         Identity& Returns the original image.
         &\\
        Autocontrast & Maximizes the image contrast by setting the darkest (lightest)
pixel to black (white). &\\
Equalize & Equalizes the image histogram.& \\
RandSolarize & Invert all pixels above a threshold value $T$.& $T \in U(0,1)$\\
RandColor & Adjust the color balance. $C=0$ returns a black\&white image, $C=1$ returns the original image. &$C \in U(0.05,0.95)$\\
RandContrast & Adjust the contrast. $C=0$ returns a solid grey image, $C=1$ returns the original image. &$C \in U(0.05,0.95)$\\
RandBrightness & Adjust the brightness. $C=0$ returns a black image, $C=1$ returns the original image. &$C \in U(0.05,0.95)$\\
RandSharpness & Adjust the sharpness. $C=0$ returns a blurred image, $C=1$ returns the original image. &$C \in U(0.05,0.95)$\\
RandPolarize & Reduce each pixel to $C$ bits. &$C \in U(4,8)$\\
\bottomrule
    \end{tabular}
}
    \caption{Appearance transformations, called \texttt{TransAppearance}, used in strong data augmentation.}
    \label{tab:TransAppearance}
\end{minipage}
\end{table}

\clearpage


\end{document}